%% file: main.tex
\documentclass[5p]{elsarticle}

\journal{Information Fusion}

\usepackage[dvipsnames]{xcolor}
\usepackage{amsmath}
\usepackage{amssymb}
\usepackage{amsthm}

\usepackage{verbatim}
\usepackage{dsfont}
\usepackage{csquotes}
\usepackage{graphicx}
\usepackage{enumitem}
\usepackage{ stmaryrd }
\usepackage{caption}
\usepackage{subcaption}
\usepackage{wrapfig}
\usepackage{booktabs}
\usepackage{multirow}
\usepackage{comment}
\usepackage{amsmath}
\usepackage{mathtools}
\usepackage{hyperref}
\usepackage{physics}
\usepackage{url}
\usepackage[export]{adjustbox}
\usepackage{appendix}

\usepackage[sectionbib]{bibunits} 

\newcommand{\x}{\boldsymbol{x}}
\newcommand{\X}{\boldsymbol{X}}
\newcommand{\y}{\boldsymbol{y}}

\newcommand{\w}{\boldsymbol{w}}
\newcommand{\ba}{\boldsymbol{a}}

\newcommand{\gi}{Gradient$\,\times\,$Input}

\newcommand{\app}[1]{\ref{#1}}

\makeatletter
\def\ps@pprintTitle{%
  \let\@oddhead\@empty
  \let\@evenhead\@empty
  \def\@oddfoot{\reset@font\hfil\thepage\hfil}
  \let\@evenfoot\@oddfoot
}
\makeatother

\defaultbibliographystyle{abbrv}
\defaultbibliography{lrpgrad}

\begin{document}

\begin{frontmatter}
\title{Preemptively Pruning Clever-Hans Strategies in Deep Neural Networks}
\author[tub,bifold]{Lorenz Linhardt}
\author[tub,bifold,korea,mpii,google]{Klaus-Robert M\"uller\texorpdfstring{\corref{cor1}}}
\author[fub,bifold,tub]{Gr\'egoire Montavon\texorpdfstring{\corref{cor1}}}

\address[tub]{Machine Learning Group, Technische Universit\"at Berlin, 10587 Berlin,
Germany}
\address[bifold]{Berlin Institute for the Foundations of Learning and Data -- BIFOLD, 10587 Berlin, Germany}
\address[korea]{Department of Artificial Intelligence, Korea University, Seoul 136-713, South Korea}
\address[mpii]{Max Planck Institute for Informatics, 66123 Saarbr\"ucken,
Germany}
\address[google]{Google DeepMind, Berlin, 
Germany}
\address[fub]{Department of Mathematics and Computer Science, Freie Universit\"at Berlin, 14195 Berlin,
Germany}
\cortext[cor1]{Corresponding authors. \\\raggedright\textit{E-mail addresses:} l.linhardt@campus.tu-berlin.de (L. Linahrdt), klaus-robert.mueller@tu-berlin.de (K.-R. M\"uller), gregoire.montavon@fu-berlin.de (G. Montavon)}

\begin{abstract}
Robustness has become an important consideration in deep learning. With the help of explainable AI, mismatches between an explained model's decision strategy and the user's domain knowledge (e.g.\ Clever Hans effects) have been identified as a starting point for improving faulty models. However, it is less clear what to do when the \textit{user and the explanation agree}.
In this paper, we demonstrate that acceptance of explanations by the user is not a guarantee for a machine learning model to be robust against Clever Hans effects, which may remain undetected. Such hidden flaws of the model can nevertheless be mitigated, and we demonstrate this by contributing a new method, Explanation-Guided Exposure Minimization (EGEM), that \textit{preemptively} prunes variations in the ML model that have not been the subject of positive explanation feedback. Experiments demonstrate that our approach leads to models that strongly reduce their reliance on hidden Clever Hans strategies, and consequently achieve higher accuracy on new data.
\end{abstract}


 \end{frontmatter}

 \begin{bibunit}

\section{Introduction}
Machine learning (ML) models such as deep neural networks have been shown to be capable of converting large datasets into highly nonlinear predictive models~\cite{Krizhevsky2012, Bahdanau2015, Mnih2015, Silver2016, Vaswani2017, Chen2020}. As ML systems are increasingly being considered for high-stakes decision making, such as autonomous driving~\cite{Aradi2022} or medical diagnosis~\cite{Capper2018, WynantsEtAl2020, RobertsEtAl2021, Jurmeister2022, Sorantin2022}, building them in a way that they reliably maintain their prediction accuracy on new data is crucial.

Proper data splitting and evaluation on hold-out test sets have long been recognized as an essential part of the validation process (e.g.\ in~\cite{Bishop2006}), but unfortunately, such techniques cannot detect all flaws of a model~\cite{Gebru2018, Lapuschkin2019, Rudin2019, Geirhos2020}. Misspecified loss functions, domain shifts, spurious correlations, or biased datasets can potentially compromise attempts to build well-generalizing models, without altering the measured accuracy.
Spurious correlations -- i.e. correlations that do not generalize -- of input variables with the label are a common threat \cite{Calude2016}: If a model learns to use a spurious correlation as part of its decision strategy, also known as the Clever Hans (CH) effect~\cite{Lapuschkin2019} or shortcut learning~\cite{Geirhos2020}, its performance will drop on data where this fake correlation ceases to hold. For example, if a bird classification model learns to recognize a species based on the background, due to a spurious correlation in the training data, it will \emph{not} be accurate when the bird is depicted in an untypical environment~\citep{SagawaKHL2020}.
In real-world scenarios, e.g.\ medical applications, a failure to address these more elusive flaws might lead to catastrophic failures, as has been demonstrated numerous times (e.g.~\cite{Winkler2019, Lapuschkin2019, hagele2020resolving, RobertsEtAl2021}). This has spurred efforts to find potential causes of such failures (e.g.~\cite{Lapuschkin2019, Wu2019, Booth2020, Tian2020}).

\begin{figure*}[!ht]
\centering
\includegraphics[width=\linewidth]{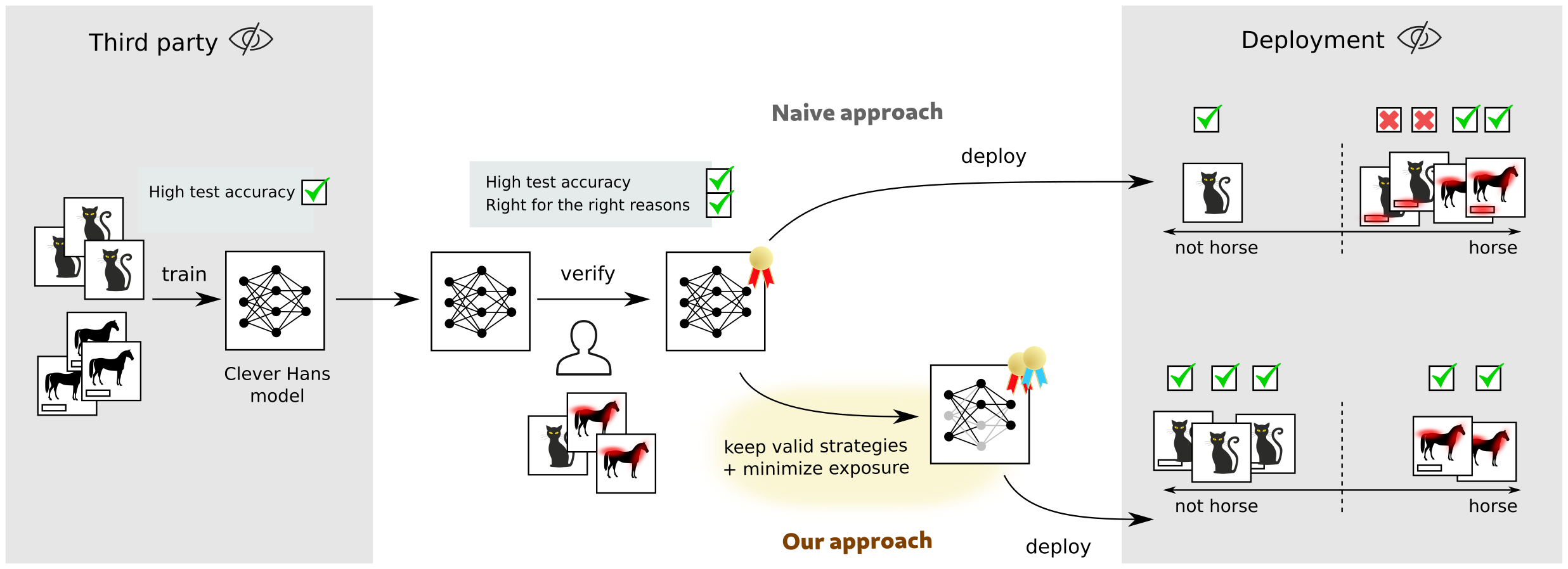}
\caption{
Comparison of a naive XAI-based validation/deployment pipeline and our proposed approach incorporating an additional exposure minimization step. \textit{Left:} A third party trains a flawed (Clever Hans) model which exploits a spurious correlation (images of the horse class have a copyright tag in the bottom-left corner). \textit{Middle:} The user receives the model. Because the user has limited data (in particular, no data with copyright tag), the flaw of the model cannot be easily detected with XAI methods and the model appears to be both accurate and right for the right reasons. \textit{Right:} Because of the undetected flaw, a naive deployment of the model is likely to result in prediction errors on new data (e.g.\ cats with copyright tags predicted to be horses). Our proposed approach preemptively reduces exposure to unseen features (the copyright tag), thereby avoiding these incorrect predictions.
}
\label{fig:intro}
\end{figure*}

Explainable AI (XAI) \cite{Gunning2019, Samek2019a, Arrieta2020,samek2021, Holzinger2022} is a natural starting point for robustification beyond classical hold-out validation because it places human experts in the loop. As demonstrated e.g.\ in \cite{Ribeiro2016, Lapuschkin2019, Plumb2021, Ghandeharioun2022}, an expert can scrutinize the model's decision strategy from the produced explanation, possibly identifying CH strategies, and remove them subsequently \cite{anders2022finding}. However, as discussed e.g.\ in~\cite{Adebayo2022}, there is no guarantee that a model that passes the XAI test can be deployed safely. Specific data points where the CH strategy reveals itself may be missing at validation time, thereby leaving the expert with the false impression that the model is free of CH effects. This is likely to happen in practice, e.g.\ when the ML practitioner is not training their own model from scratch but relies instead on a model trained by a third-party (e.g.\ \textit{foundation models}\footnote{Foundation models~\cite{Bommasani2021} are pretrained multi-purpose models typically made available by a third party (e.g.~\citep{Devlin2019, Brown2020, Oquab2023}).}) and does not have full access to the training data.

In this paper, we tackle for the first time the problem of undetected CH strategies (i.e.\ the case where the model remains flawed in spite of a full agreement of the human with the model's predictions and the generated explanations). We contribute a novel algorithm, called Explanation-Guided Exposure Minimization (EGEM), which distills from the original model a \textit{refined model} with lower overall exposure to input features but that also preserves the few valid prediction strategies contained in the generated explanations.
With mild approximations, the optimization problem embodied in our EGEM approach reduces to a simple soft-pruning rule, which can be easily implemented in a broad range of neural network architectures including convolutional networks and transformers. Our proposal, as well as the context in which it operates, are illustrated in Fig.\ \ref{fig:intro}.

To evaluate our approach, we simulate several scenarios of a user receiving a third-party model and possessing a subset of the data on which no CH strategies can be detected by classical XAI pipelines (e.g.\ LRP/SpRAy \cite{Bach2015, Lapuschkin2019}). Results on image and text data demonstrate that our proposed EGEM approach (and our extension PCA-EGEM) delivers models with a much lower reliance on CH strategies, thereby achieving more stable prediction accuracy, especially when considering data with spurious features. Our approach furthermore outperforms a number of existing and contributed baselines.

\section{Related Work}
\label{sec:related work}
In this section, we present related work on validating ML models that goes beyond classical validation techniques such as holdout or cross-validation \cite{Bishop2006} in order to address statistical artifacts such as domain shifts and spurious correlations. We make a distinction between methods relying on Explainable AI and users' explanatory feedback (Section \ref{section:xai}), and a broader set of methods addressing domain shift and spurious correlations by statistical means (Section \ref{section:robustness}).

\subsection{Explainable AI and Clever Hans}
\label{section:xai}
Explainable AI (XAI)~\cite{DBLP:journals/aim/GunningA19, Mittelstadt2019, Samek2019a, Zhang2021, samek2021, Holzinger2022} has been a major development in machine learning which has enabled insights into a broad range of black-box ML models. It has shown to be successful at explaining complex state-of-the-art neural networks classifiers \cite{Bach2015, Ribeiro2016, DBLP:journals/ijcv/SelvarajuCDVPB20, Zhang2021-hierarchical, Jeon2022, Schnake2022}, as well as regressors~\citep{Letzgus2022} and a broader set of ML techniques such as unsupervised learning (e.g.\ \cite{Eberle2020, Liznerski2020,Kauffmann2022}). While most XAI methods generate an explanation for individual instances, solutions have been proposed to aggregate them into dataset-wide explanations that can be concisely delivered to the user \cite{Lapuschkin2019}. In contrast to these works, we are not concerned with improving XAI techniques themselves, e.g.\ the interpretability of explanations, but rather with using these techniques to make models more robust against spurious correlations.

More closely related to our aims, several techniques have applied XAI for the purpose of revealing CH features in ML models~\cite{Lapuschkin2019, Plumb2021, Ghandeharioun2022, Wu2023}. Knowledge about the CH features can be used to desensitize the model to these features (e.g.\ via retraining~\cite{Plumb2021} or layer-specific adaptations~\cite{anders2022finding}). If ground-truth explanations are available (e.g.\ provided by a human expert), the model may be regularized to match these explanations~\cite{Ross2017, Rieger2019, Simpson2019}, e.g.\ by minimizing the error on the explanation via gradient descent. Such adaptations to the users' expectations have also been shown to be effective in interactive settings~\cite{Teso2019, Schramowski2020}. Our approach differs from these works as we address the case where the available data does not contain CH features, hence making them indiscoverable by the techniques above.

A further approach is DORA~\cite{Bykov2022}, which attempts to find potential CH features in a data-agnostic way and subsequently uses the discovered candidate features to detect faulty decision strategies at deployment time by identifying samples that are outliers in the DNN's internal representation space. In contrast, we attempt to \textit{robustify} the network with no need for further post-processing at deployment. Furthermore, we examine the scenario where a limited amount of clean data is available, allowing us to employ conceptually different criteria besides outlierness.

\subsection{Robustness to Spurious Correlations}
\label{section:robustness}
Our work is part of a larger body of literature concerned with \textit{domain} or \textit{covariate shift} and how to design models robust to it. Yet, it concerns itself with a decidedly specialized and rather recent part of this area: unlearning or avoiding the use of spurious features in deep neural networks.
Previous work attempting to create models that are robust against spurious correlations approached the problem from the angle of optimizing worst-group loss~\cite{Creager2021, Hu2018, SagawaKHL2020, SagawaRKP2020, Sohoni2021, Idrissi2022, Kirichenko2022, Nam2022, Fremont2019}. This approach has been shown to be effective in reducing reliance on CH features. Yet, these methods require access to samples containing the CH features and a labeling of groups in the data induced by these features. In particular, as previously pointed out by Kirichenko et al.~\cite{Kirichenko2022}, Group-DRO (distributionally robust optimization)~\cite{Hu2018}, subsampling approaches~\cite{SagawaRKP2020, Idrissi2022} and DFR (deep feature reweighting)~\cite{Kirichenko2022} assume group labels on the training or validation data, and even methods that do away with these assumptions need to rely on group labels for hyper-parameter tuning~\cite{LiuEtAl2021, Creager2021, Idrissi2022}. 
Our setting is different from the ones above in that we assume that a pretrained model is to be robustified post hoc with limited data and that data from the groups containing the CH feature are not available at all. We believe this is a highly relevant scenario, considering the increasing prevalence of pretrained third-party models that have been trained on datasets that are unavailable or too large to fully characterize.

\section{Explanation-Guided Exposure Minimization (EGEM)}
\label{section:em}
Before presenting our main technical contribution, let us restate the key aspects of the application scheme studied in this paper (as depicted in Fig.\ \ref{fig:intro}):
\begin{enumerate}[label=\roman*)]
\itemsep0em 
    \item a pretrained model provided by a third party and potentially affected by a CH effect,
    \item the unavailability of the original training data to the user, which prevents the discovery of CH features,
    \item limited data which is available to the user to validate and refine the third-party model, and for which 
    \item the conclusion of the user is that the predictions and the associated decision strategies (as revealed by XAI) are correct. I.e.\ the data available to the user is free of CH features.
\end{enumerate} 
As argued before, in spite of the positive XAI-based validation outcome, there is no guarantee that the model's decision strategy remains correct in regions of the input space not covered by the available data (e.g.\ where the CH artifact could be expressed).

As a solution to the scenario above, we propose a preemptive model refinement approach, which we call \textit{Explanation-Guided Exposure Minimization (EGEM)}. Technically, our approach is a particular form of knowledge distillation where the refined (or distilled) model should reproduce observed prediction strategies (i.e.\ predictions and explanations, hence  \textit{``explanation-guided''}) of the original model on the available data. At the same time, the refined model should \textit{minimize} its overall sensitivity or \textit{exposure} to variations in the input domain so that uninspected (potentially flawed) decision strategies are not incorporated into the overall decision strategy.

Let the original and refined model have the same architecture but distinct parameters $\theta_\text{old}$ and $\theta$. We denote by $f(\x,\theta_\text{old})$ and $f(\x,\theta)$ the predictions produced by the two models, and the explanations associated to their predictions as $\mathcal{R}(\x,\theta_\text{old})$ and $\mathcal{R}(\x,\theta)$ respectively.
We then define the learning objective as
\begin{align}
\min_\theta ~ \mathop{\mathbb{E}}\big[ \|\mathcal{R}(\x, \theta) - \mathcal{R}(\x, \theta_\mathrm{old})\|^2 + \Omega_{\x}(\theta) \big]
\label{eq:objective}
\end{align}
where the expectation is computed over the available data, and where $\Omega_{\x}(\theta)$ is a function that quantifies the exposure of the model to the input variation in general, or in some neighborhood of $\x$. In the latter case, the neighborhood should be large enough to encompass instances outside the available data such as those encountered at deployment time.

Although the formulation of Eq.\ \eqref{eq:objective} is general, it is not practical, because it would require optimizing a highly nonlinear and non-convex objective. Moreover, the objective depends on explanation functions and on a complexity term that themselves may depend on multiple model evaluations, thereby making the optimization procedure intractable.

\begin{figure*}[!ht]
\centering
\includegraphics[width=\linewidth]{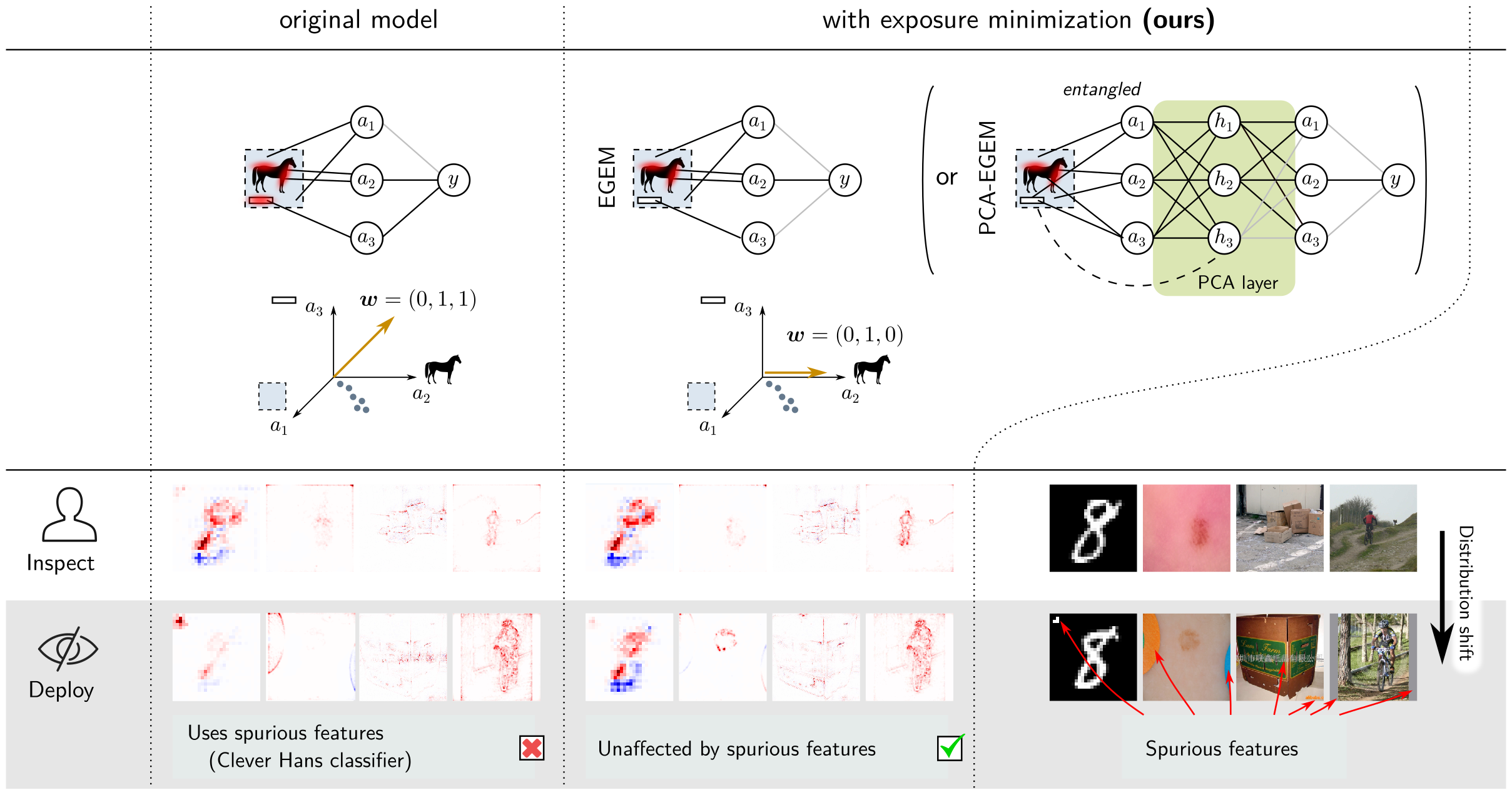}
\caption{\textit{Top:} Cartoon depicting the removal of unseen Clever Hans strategies via our proposed exposure minimization approaches (EGEM and PCA-EGEM). The refined model only retains the dependence on $a_2$ (a neuron detecting the actual horse) and removes its reliance on $a_3$ (a neuron responsive to spurious copyright tags). \textit{Bottom:} Qualitative behavior of PCA-EGEM on ML models trained on real datasets. The models produced by our approach become robust to spurious features not seen at inspection time but occurring at deployment time. (Pixel-wise explanations are computed using the \texttt{zennit} package~\cite{zennit}.)}
\label{fig:explanation-refine}
\end{figure*}

\subsection{A Practical Formulation for EGEM}
\label{section:em:deep}
To make the concept of explanation-guided exposure minimization effective, we will restrict the class of XAI methods on which it depends to those (1) that can attribute onto any layer of the model and (2) whose produced scores have the following properties: the score assigned to a neuron $i$ at a given layer should be decomposable in terms of the neurons $j$ in the layer above, i.e.\ $R_i = \sum_{j} R_{ij}$ and terms of the decomposition should have the structure
\begin{align}\textstyle
R_{ij} = a_i \, \rho(w_{ij}) \, d_j
\label{eq:struct}
\end{align}
where $a_i$ denotes the activation of neuron $i$, $w_{ij}$ is the weight connecting neuron $i$ to neuron $j$ in the next layer, $\rho$ is an increasing function satisfying $\rho(0)=0$ (e.g.\ the identity function), and $d_j$ is a term that only indirectly depends on the input activations and parameters in the given layer and that is reasonable to approximate as constant locally. Explanation techniques that produce explanation scores with such structure include propagation methods such as Layer-wise Relevance Propagation (LRP)~\cite{Bach2015, Montavon2019a} or its limit \gi{} (GI)~\citep{Shrikumar2017} and Integrated Gradients (IG)~\citep{Sundararajan2017}. (See~\app{app:relevance-structure} for derivations.)

An advantage of expressing explanations in the form of Eq.\ \eqref{eq:struct} is that it gives access to higher-level abstractions (e.g.\ visual concepts) built by the network, which are often represented more naturally in these layers. If we further restrict the search for refined parameters to the weights of the same layer, one can then formulate an alternative to the optimization problem of Eq.\ \eqref{eq:objective} which is more expressive and also computationally more tractable:
\begin{align}
\min_w ~ \sum_{ij} \mathbb{E}  \Big[ \big(\overbrace{\big.a_j \rho(w_{ij}) d_j}^{\displaystyle R_{ij}} &- \overbrace{\big.a_i \rho(w_{ij}^\mathrm{old}) d_j}^{\displaystyle R_{ij}^\mathrm{old}}\big)^2\nonumber\\[-3mm]
&+\underbrace{\lambda \cdot \big(\big.\rho(w_{ij}) d_j\big)^2}_{\displaystyle \Omega_{ij}} \Big]
\label{eq:objective-layerwise}
\end{align}
where, as for Eq.\ \eqref{eq:objective}, the expectation is taken over the available data. Here, $w_{ij}^\text{old}$ and $w_{ij}$ denote the original and the refined weights.
As in Eq.\ \eqref{eq:objective}, the first squared term guides the explanations of the refined model to be close to that of the original model and the second squared term ($\Omega_{ij}$) carries out the exposure minimization. The latter can be interpreted as the extent to which the refined model responds to the activation of neuron $i$ through neuron $j$, in particular, if it becomes zero, the model becomes unresponsive to the activation of neuron $i$. The lack of direct dependence of $\Omega_{ij}$ on the input activations is instrumental for it to generalize outside the available data towards data encountered at deployment time. An advantage of the formulation of Eq.\ \eqref{eq:objective-layerwise} is that it has the closed-form solution:
\begin{align}
\forall_{ij}:~w_{ij} = \frac{\mathbb{E}[a_i^2d_j^2]}{\mathbb{E}[a_i^2d_j^2] + \lambda \mathbb{E}[d_j^2]} w_{ij}^\text{old}
\label{eq:weights-layerwise}
\end{align}
See~\app{app:egem:derivation} for a derivation.
In other words, the refined model can be seen as a soft-pruned version of the original model where the pruning strength depends on how frequently and to what magnitude the input neuron is activated and how the model responds to the output neuron.

If we further assume that $a_i$ and $d_j$ are independent, then $d_j$ vanishes from Equation \eqref{eq:weights-layerwise}, leading to the simpler pruning rule
\begin{align}
\forall_{ij}:~w_{ij} = \frac{\mathbb{E}[a_i^2]}{\mathbb{E}[a_i^2] + \lambda}w_{ij}^\text{old}.
\label{eq:pruning}
\end{align}
Lastly, because the pruning coefficients only depend on the input neuron, the same pruning can be achieved by keeping the weights intact and inserting a layer directly after the input activations that performs the scaling:
\begin{align}
\forall_i:~a_i \gets a_i c_i
\label{eq:activations-layerwise}
\end{align}
with $c_i = \mathbb{E}[a_i^2] / (\mathbb{E}[a_i^2] + \lambda)$. 
The pruning of the neural network architecture and the resulting loss of dependence on the CH feature are depicted in Fig.\ \ref{fig:explanation-refine} (top). Due to the soft character of the pruning, i.e. $c_i\in [0,1] $, EGEM is able to make more subtle changes to the network than explanation-based hard-pruning approaches, such as the one by Yeom et al.~\citep{Yeom2021}.

The same Eq.\ \eqref{eq:activations-layerwise} can also be applied to convolutional layers. To calculate the scaling parameters $c_i$, the activations of each channel are summed up along the spatial dimensions. For refinement, the pruning coefficients are then applied to all activations of the corresponding feature map (cf.\ Eq.\ \ref{eq:activations-layerwise}). Such pruning strategy for convolutional layers can be derived exactly from Eq.\ \eqref{eq:objective-layerwise} if assuming activation maps of infinite size (or circular convolutions) and stride $1$. For the majority of convolution layers used in practice, Eq.\ \eqref{eq:activations-layerwise} only derives from the objective formulation approximately.

\subsection{Pruning in PCA Space}
\label{section:em:pcaegem}
Within the EGEM soft-pruning strategy, each dimension of a layer is pruned individually. In practice, this only allows to eliminate undetected flawed strategies that use a set of neurons that is disjoint from neurons supporting the validated strategies. Because a given neuron may in practice contribute to both, the standard version of EGEM would not be able to carry out the exposure minimization task optimally.
To address this limitation, we propose PCA-EGEM, which inserts a virtual layer, mapping activations to the PCA space (computed from the available data) and back (cf.\ Fig.\ \ref{fig:explanation-refine}).
PCA-EGEM then applies soft-pruning as in Eq.~\eqref{eq:activations-layerwise}, but in PCA space, that is:
\begin{align}
    h_k &= U_k^\top (\ba - \bar{\ba}) \nonumber\\
    h_k &\gets h_k c_k
    \label{eq:pcaegem}\\
    \ba &\gets \textstyle \sum_k U_k h_k + \bar{\ba} \nonumber
\end{align}
with $c_k = \mathbb{E}[h_k^2] / (\mathbb{E}[h_k^2] + \lambda)$.
Here, $\{U_k\}_{k=1}^K$ is the basis of PCA eigenvectors and $\boldsymbol{\bar{a}}$ is the mean of the activations over the available data. The motivation for such mapping to the PCA space is that activation patterns that support observed strategies will be represented in the top PCA components. PCA-EGEM can therefore separate them better from the unobserved strategies that are likely not spanned by the top principal components.

While using PCA to find principal directions of interpretable features has been proposed previously by Härkönen et al.~\cite{Harkonen2020} for GANs~\cite{Goodfellow2014} and has found application beyond that~\cite{Zhou2021, Chormai2022}, to our knowledge its use for the purpose of identifying a basis for exposure minimization is novel.

\section{Experimental Evaluation}
\label{section:eval}
In this section, we evaluate the efficacy of the approaches introduced in Section~\ref{section:em} on datasets that either naturally contain spurious correlations, giving rise to Clever Hans decision strategies, or have been modified to introduce such correlations. Each experiment involves three datasets for which we ensure the following structure: 
\begin{enumerate}[label=\roman*)]
\itemsep0em 
    \item the \textbf{training data}, used for training the original model, containing CH features, and resulting in a model with a CH decision strategy. The training data is for its larger part not available to the user.
    \item the \textbf{available data}, much smaller than the training data, and \emph{not} containing any CH feature. It is the only data available to the user for validating and refining the model.
    \item the \textbf{test data}, disjoint from i) and ii), and used to evaluate model refinement strategies. It is poisoned to contain the CH feature to 0\% or 100\% on all classes. Unlike in the training data, the CH feature is thus decorrelated from the target class.
\end{enumerate}

After introducing the datasets, we demonstrate that the proposed approaches can mitigate the CH effect learned by pretrained models.
Additionally, we empirically explore the effect of the number of samples used for refinement and discuss the challenges of hyper-parameter selection in Sections~\ref{sec:eval:hyper} and~\ref{section:eval:samples}, as well as the impact of the type of CH feature in Section~\ref{section:eval:mnist}. A more qualitative evaluation on the CelebA dataset~\cite{LiuLWT2015} follows in Section~\ref{section:eval:celeba} and an application to sentiment classification is described in Section~\ref{section:eval:movies}.

\subsection{Datasets}
\label{section:data}
We introduce here the datasets used to evaluate the proposed methods in Sections~\ref{section:eval:accuracy}-\ref{section:eval:samples}: a modified version of the MNIST dataset~\cite{mnist}, the ImageNet dataset~\cite{Deng2009,Russakovsky2015}, and the ISIC dataset~\cite{Codella2018, Tschandl2018, Combalia2019}.
Details on the preprocessing and the neural networks used for each dataset can be found in~\app{app:train} and~\app{app:experiments}. Dataset statistics can be found in Table~\ref{tab:experiments:acc_overview}.
\begin{table*}[tb]
\centering
\footnotesize
\setlength{\tabcolsep}{4pt}
\begin{tabular}{lrllrrrr}
\toprule
Dataset & \# classes & class names & CH feature & $p(\text{CH})$ & $N_{\text{train}}$  & $N_{\text{available}}$  & $N_{\text{test}}$ 
   \\ \midrule
MNIST & 10 & \{0, 1, 2, 3, 4, 5, 6, 7, \textbf{8}, 9\} & 3-pixel corner & 0.700 & 60,000 & 700 & 10,000
   \\ 
ISIC2019 & 8 & \{\textbf{Melanocytic nevus}, \dots \} &
  colored patch & 0.015 & 22,797 & 700 & 2,534
   \\ 
ImageNet subtask 1 & 2 & \{\textbf{carton}, crate\} &
  watermark, www & 0.438 & 1.2M & 700 & 100
   \\ 
ImageNet subtask 2 & 2 & \{\textbf{carton}, envelope\} &
  watermark, www & 0.438 & 1.2M & 700 & 100
   \\ 
ImageNet subtask 3 & 2 & \{\textbf{carton}, packet\} &
  watermark, www & 0.438 & 1.2M & 700 & 100
   \\ 
ImageNet subtask 4 & 2 &\{\textbf{mountain bike}, bicycle-built-for-two\} &
  gray frame & 0.028 & 1.2M & 700 & 100
   \\ \bottomrule
\end{tabular}
\caption{Overview of datasets, their class structure (class with CH feature in \textbf{bold}), a description of the CH feature, and the number of instances in each set of data.
p(CH) is the fraction of training samples of the affected class containing the CH feature.}

\label{tab:experiments:acc_overview}
\end{table*}
\paragraph{Modified MNIST} We create a variant of the original MNIST dataset~\cite{mnist} in which digits of the class `8' are superimposed with a small artifact in the top-left corner (see Fig.~\ref{fig:explanation-refine}). In order to generate a natural yet biased split of the training data that separates an artifact-free set of data available for refinement, we train a variational autoencoder~\cite{Kingma2014} on this modified dataset and manually chose a threshold along a latent dimension such that the samples affected with the artifact only fall on one side. This defines a subset of 39,942 samples from which clean datasets are sampled for refinement and leaves a systematically biased subset (containing all modified `8' samples) only accessible during training. We train a neural network (2 convolutional and 2 fully connected layers) on this dataset using binary cross-entropy loss over all ten classes on the whole training data.

\paragraph{ISIC} The ISIC 2019 dataset~\cite{Codella2018, Tschandl2018, Combalia2019} consists of images containing skin lesions that are associated with one of eight medical diagnoses. We fine-tune a neural network based on a VGG-16 pretrained on ImageNet for this classification task using a cross-entropy loss. Some images of the class `Melanocytic nevus' are contaminated with colored patches (see Fig.~\ref{fig:explanation-refine}), which have been recognized as potential CH feature~\cite{Mishra2016, Rieger2019, Bissoto2020, anders2022finding}. We manually remove all contaminated images after training and use this clean dataset for refinement. Images in the test set are contaminated at the desired ratio by pasting one extracted colored patch onto other images. 

\paragraph{ImageNet} We use the ILSVRC 2012 subset of the ImageNet dataset~\cite{Deng2009,Russakovsky2015}. Previous work has identified multiple spurious correlations potentially affecting a model's output~\cite{anders2022finding}. From these known spurious features we select two that lead to an easily reproducible CH effect in popular pretrained models. In particular, we use a watermark and web-address on images of the `carton' class and a gray frame around images of the `mountain bike' class as CH features (see Fig.~\ref{fig:explanation-refine}). We vary their frequency in the test set by pasting these features on images (details in~\app{app:data}). The selected classes are evaluated in a binary classification setting against the most similar classes in terms of the output probabilities. Training set images used for refinement that do not contain the CH feature are manually selected for the `carton' experiments and automatically for `mountain bike' experiment.
For experiments on this dataset, we make use of the pretrained ResNet50~\cite{He2016} (for the `carton' class) and VGG-16~\cite{Simonyan2015} (for the `mountain bike' class) networks available in pytorch\footnote{\url{www.pytorch.org}~\cite{Paszke2019}}. 
\subsection{Methods}
\label{section:eval:methods}
We evaluate both EGEM and PCA-EGEM and compare them to several baseline methods for the mitigation of the Clever Hans effect. 
\begin{enumerate}[label=\roman*)]
    \item \textbf{Original:} This is the original model without any modifications. Note that approaches for mitigating the Clever Hans effect, such as~\citep{Plumb2021, anders2022finding} reduce technically to this simple baseline because they only modify the model in presence of \textit{detected} CH features, whereas in our scenario, no CH features are detectable in the available data.
    \item \textbf{RGEM:} This baseline, which we contribute, is a modification of EGEM, where the exposure minimization is carried out under the constraint of preserving model \textit{response} instead of model \textit{explanation}. Specifically, RGEM optimizes the objective:
\begin{align}
    \min_{\theta} ~~ \mathbb{E}[(f(\x, \theta) - f(\x, \theta_\text{old}))^2] + \lambda \|\theta\|^2
    \label{eq:rgem}
\end{align}
    with $f$ being the neural network function and $\theta$ the parameters in the last layer. Compared to EGEM, RGEM can only refine the last layer weights.
    \item \textbf{Ridge:} This baseline consists of replacing the last layer weights of the original model by weights learned via ridge regression on the available data. It is equivalent to linear probing or deep feature reweighting~\cite{Kirichenko2022}, which has been shown to be effective in mitigating accuracy loss due to subpopulation shifts~\cite{Santurkar2020} and the Clever Hans effect when hyper-parameter selection based on worst-group accuracy optimization is possible~\cite{Kirichenko2022}. The formulation for Ridge can also be retrieved by replacing the output of the original model, $f(\x, \theta_\text{old})$, in the formulation of RGEM (see~\app{app:rgem}) with the ground-truth labels.
    \item \textbf{Retrain:} This baseline corresponds to a deeper retraining, where starting from the original model, layers are fine-tuned to improve the classification of the available data. Unlike the RGEM and Ridge baselines, all layers are updated to fit the available data. This makes fine-tuning the most flexible but also computationally costly approach of all the compared methods.
\end{enumerate}

\subsection{Evaluation Setup}
\label{section:eval:setup}
We evaluate all methods along two dimensions: in-distribution accuracy, which is measured as classification accuracy on CH-free (clean) data, and robustness against the CH effect. To evaluate the latter, we generate artificially a fully poisoned test set by adding the CH artifact to all test images of all classes. Artificially poisoned data allows us to isolate the CH effect, as it is identical to the clean test data, except for the CH feature. This setup allows us to attribute any difference in accuracy between these two (clean and manipulated) test sets to the presence of the CH feature. In this scenario, the correlation of the CH feature and the target class breaks. Such a distribution shift could, for example, happen in medical applications where a classifier might be trained on data in which the mode of data collection or the population characteristics of subjects are correlated with the outcome, but where this spurious correlation does not hold in the general case~\cite{RobertsEtAl2021}. Note that while this poisoning scenario is an extreme case, it is not the worst case, as the class that was contaminated during training will also be modified with artifacts during testing.

For refinement, 700 correctly predicted instances per class are used, oversampling images if fewer than 700 correctly predicted samples are in the available data to maintain a balance between classes and ensure that all classes have equal contribution to the refinement processes. For the modified MNIST and the ISIC dataset, we use 1000 randomly chosen test samples for each run of the evaluation, for ImageNet we use all available validation samples.
We evaluate the various models for all tasks under 0\% and 100\% uniform poisoning. Classification accuracy for intermediate levels of poisoning can be obtained by linear interpolation of these extremes. 
\subsection{Results}
\label{section:eval:accuracy}
\begin{figure*}[htb!]
    \centering
    \includegraphics[width=\textwidth]{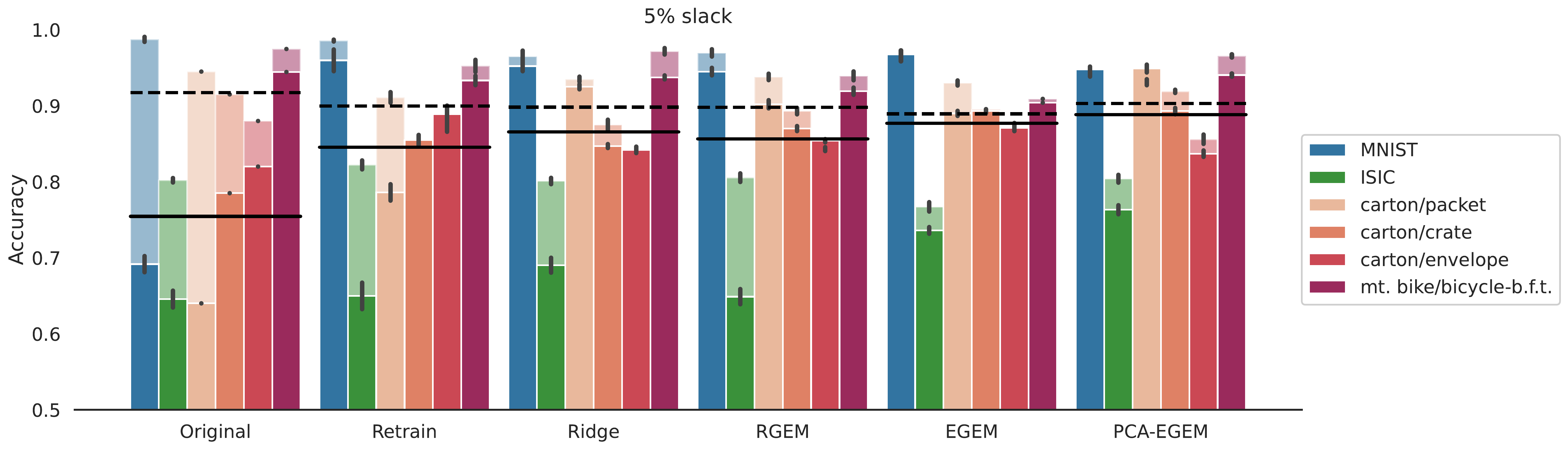}
    \caption{The accuracy for 0\% (lighter shade) and uniform 100\% (darker shade) poisoning with the spurious feature. The last four bars for each method refer to the binary tasks constructed from the ImageNet dataset. Solid lines show average 100\%-poisoned accuracy and dashed lines show average clean-data accuracy over all datasets. The results shown are the mean accuracy and standard deviation obtained using 700 refinement samples per class on the respective test sets over five runs.}
    \label{fig:eval:pois_bars}
\end{figure*}
Figure~\ref{fig:eval:pois_bars} shows the obtained accuracy under 0\% and 100\% poisoning.
An ideal model would obtain high accuracy with only a very small difference between clean and poisoned data. It should be invariant to the spurious feature and at most react to possible interference with other features, e.g.\ the spurious feature being pasted on top of a relevant part of the image, while not losing accuracy on the clean data. As expected, across all datasets increased poisoning reduces the accuracy of the original model. \textit{Importantly, this drop in accuracy cannot be detected without access to samples containing the CH feature.} 

\paragraph{Modified MNIST} On the modified MNIST dataset, the original model loses about 30\% of its clean-data accuracy when evaluated at the 100\% poisoning level. All other models achieve both clean-data and 100\%-poisoned accuracy levels within 4\% of the original model's clean-data accuracy. While explanation-based methods lose slightly more clean-data accuracy than the other baselines, they display virtually no gap between clean-data accuracy and 100\%-poisoned accuracy, making them the most predictable when no poisoned data is available.

\paragraph{ISIC} On the more complex ISIC dataset, it can be observed that exposure to the CH feature cannot be completely removed by any of the methods. EGEM and PCA-EGEM still provide fairly robust models, with the highest poisoned-data accuracy and the smallest gap between 0\%-poisoned and 100\%-poisoned accuracy.
PCA-EGEM retains clean-data accuracy while being the only method improving poisoned-data accuracy by more than 10 percentage points. The dataset provides a challenge for all other methods. Even though Retrain is the only method that improves clean-data accuracy in the refinement process, its poisoned-data accuracy is virtually the same as the original model's, indicating that the absence of a feature in the available data is not enough to remove it from the model, given only a limited amount of samples.

\paragraph{ImageNet} On the ImageNet tasks containing the `carton' class, PCA-EGEM is the most robust refinement method. It is only outperformed in the 100\%-poisoned setting of the `carton/envelope' task, where Retrain achieves the highest clean-data and poisoned accuracy. As we will see in Section~\ref{sec:eval:hyper}, the inferior 100\% poisoning accuracy of PCA-EGEM is a result of the hyper-parameter selection procedure and not fundamentally due to the pruning-based nature of the method. 
On the 100\%-poisoned setting `mountain bike' task, no refinement method is able to achieve accuracy gains over the original model. This might be due to the small magnitude of the CH effect resulting in the clean-data loss due to refinement outweighing the robustness gain. This case also demonstrates that refinement is not beneficial in all scenarios and might not even lead to an improved 100\%-poisoned accuracy. Whether or not to refine should be decided based on whether the loss of clean-data accuracy can be tolerated.

\medskip \noindent We hypothesize that the reason for Ridge and RGEM being suboptimal for some of the datasets is that these approaches can only manipulate the last layer of the network, which may not be where the CH feature is expressed most clearly~\citep{anders2022finding, Bykov2022, Kauffmann2022}. An analysis of the per-layer separability of clean and manipulated images supports this by showing that in particular for the ISIC dataset and the `mountain bike' task, separability is highest at layers close to the input (see~\app{app:exp:misc}).

Another concern that should be taken into consideration when applying Ridge, Retrain, or RGEM is that those approaches may learn new spurious correlations on the limited available data. While we separated the available data into 80\% training and 20\% validation splits, newly emerging CH effects may still go unnoticed. Since (PCA-)EGEM can only \textit{reduce} the sensitivity of the network, it precludes the learning of new, potentially CH, features. 

Overall, the results in this section demonstrate that the proposed refinement methods can preemptively robustify a pretrained model against Clever Hans effects, even if the latter cannot be observed from the limited available data. We could clearly establish that the attempt to robustify against CH behavior in absence of the associated artifact or knowledge thereof is not a hopeless endeavor and can be addressed with relatively simple methods. Yet, the trade-off between clean-data accuracy and poisoned-data accuracy cannot be directly observed on real data and thus needs to be resolved heuristically. We explore this aspect in the next section.

\subsection{Hyper-parameter Selection} 
\label{sec:eval:hyper}
\begin{figure*}[htb!]
    \centering
    \includegraphics[width=\textwidth]{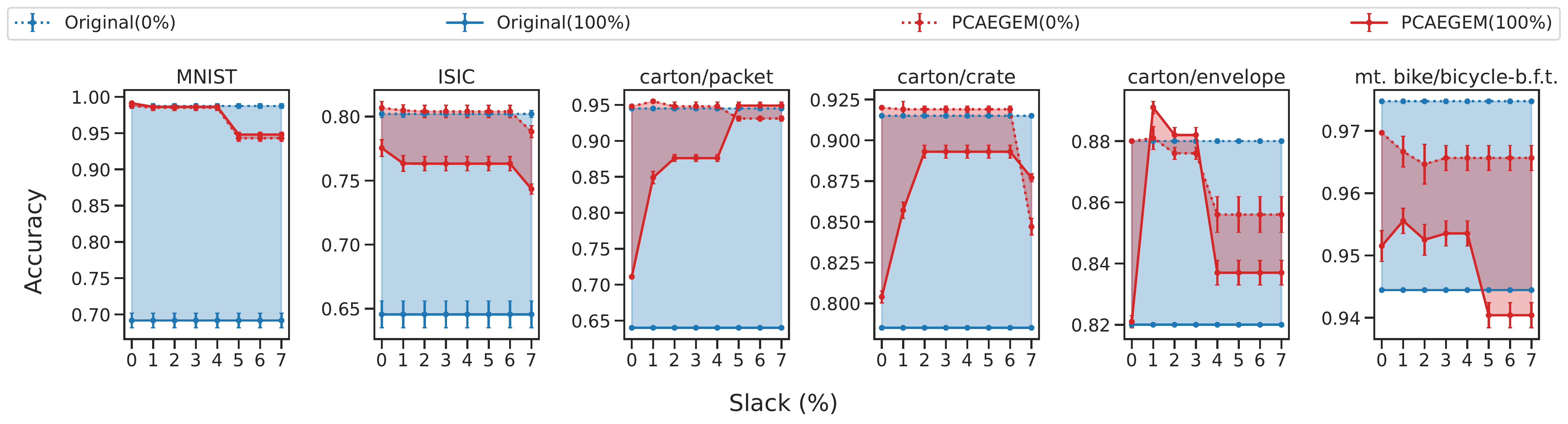}
    \caption{Accuracy under variations of the slack parameter. Higher slack means higher refinement-data loss is accepted when selecting the refinement hyper-parameter. The dotted line indicates clean-data accuracy whereas the solid line indicates 100\%-poisoned data accuracy. Mean and standard deviation are computed over 5 runs.}
    \label{fig:eval:slack}
\end{figure*}
The hyper-parameters optimized in the experiments in this section are the number or epochs for `Retrain' and the regularization factor $\lambda$ for all other refinement methods. 
For the deep exposure-based approaches, EGEM and PCA-EGEM, we do not optimize $\lambda$ for each layer directly, but we rather employ an approach inspired by the triangular method of Ashouri et al.~\cite{Ashouri2019} and earlier work~\cite{Polyak2015} where pruning strength increases with the layer index, which allows us to reduce the number of parameters to optimize to one. In particular, we define thresholds $\tau_l$ that denote the desired average pruning factor per layer, where $l$ is the layer index within the set of $L$ layers to be refined:
\begin{align*}
\tau_l = 1 - (1-\alpha) \times \frac{l-1}{L-1}
\end{align*}
and optimize $\alpha$. $\lambda_l$ is then set such that the average pruning factor $\mathop{\mathbb{E}}_{\x\in\X}\mathop{\mathbb{E}}_{j}c_j$ from Eq.~\eqref{eq:weights-layerwise} for layer $l$ is at most $\tau_l$. The search for $\lambda_l$ given $\tau_l$ can be easily implemented as exponential search. 

Ideally, the hyper-parameters should be set such that classification loss is minimized while exposure to the spurious artifact is negligible. While classification loss on clean data can be readily approximated by evaluating the loss function on the available data, exposure to the spurious artifact is a more elusive quantity and cannot be measured without prior knowledge of the spurious artifact. 
In previous work (e.g.~\cite{Hu2018, SagawaKHL2020, SagawaRKP2020, Creager2021, Sohoni2021, Idrissi2022, Kirichenko2022, Nam2022}) it is assumed that for each class a set of samples with and without the spurious artifact is given and in most cases that the \textit{worst-group-accuracy} can be directly optimized or at least used for hyper-parameter selection, circumventing this problem. Since in our problem setting access to samples with the artifact is not given, this metric for parameter selection is not available and we need to establish a heuristic approach.
Assuming that the classification loss on clean data can be approximated accurately, one option is to pick the strongest refinement hyper-parameter (i.e.\ highest number of epochs, largest $\lambda$ or smallest $\alpha$) from a pre-defined set (see~\app{app:experiments}) for which the validation accuracy after refinement is at least as high as the one achieved by the original model. 

As it is possible that strong refinement also impairs the use of generalizing features, there may be a trade-off between clean-data accuracy and robustness to spurious features. That optimizing overall clean-data accuracy is generally not the best approach to optimizing overall accuracy is highlighted by the fact that other works optimize \textit{worst-group-accuracy}, as mentioned above. We explore the accuracy trade-off in Fig.~\ref{fig:eval:slack} by introducing a `slack' parameter $s$ to the hyper-parameter selection for PCA-EGEM. We refer to~\app{app:slack} for the results of all methods. The refinement hyper-parameter is then chosen as the strongest regularization, given that the validation accuracy is at most $s$\% smaller than the one achieved by the original model. The idea is that minimizing loss of classification accuracy on the available data prevents removing too much exposure to useful features, yet, allowing for some slack counteracts the tendency to choose trivial least-refinement solutions. 

We suspect that in the simple case of the modified MNIST dataset, the model only learned few important high-level features and that the CH feature is close to being disentangled in some layer of the network. This scenario is a natural fit for pruning methods which could simply remove the outgoing connections of the node corresponding to the CH feature. Stronger refinement risks pruning useful features as well, which is an effect that can be observed in Fig.~\ref{fig:eval:slack}. 
For most datasets, we can observe that the accuracy curves first converge to or maintain a minimal 0\%-100\% poisoning gap. In this regime, PCA-EGEM prunes unused or CH features. After crossing a certain level of slack, both accuracy values deteriorate as features necessary for correct classifications are being pruned as well. 

The results previously presented in Fig.~\ref{fig:eval:pois_bars} are the outcomes for $s=5$\%. This is a heuristic and it can be seen from Fig.~\ref{fig:eval:slack} that different values of slack may be beneficial to increase robustness, depending on the dataset. We also show in~\app{app:slack} and~\app{app:samples} that PCA-EGEM provides the most robust refinement over a large range of slack values. As slack cannot be optimized w.r.t.\ the true deployment-time accuracy, we propose to set $s$ between 1\% and 5\% as a rule of thumb based on these experiments, where smaller data samples may permit lower slack values.

In principle, another hyper-parameter is the choice of layers to refine. Knowledge of the type of Clever Hans could potentially guide this choice~\cite {anders2022finding, Lee2022} as the layer in which a concept is best represented may differ across concepts~\cite{Kim2018}. Since we do not assume such knowledge in our experiments, we simply refine the activations after every ResNet50 or VGG-16 block for the parts of the models that are derived from those architectures and additionally after every ReLU following a fully connected or convolutional layer that is not contained in a ResNet or VGG block. For `Retrain' we fine-tune the whole network and RGEM and Ridge are restricted to the last layer.

\subsection{The Effect of the Sample Size}
\label{section:eval:samples}
\begin{figure*}[tb!]
    \centering
    \includegraphics[width=\textwidth]{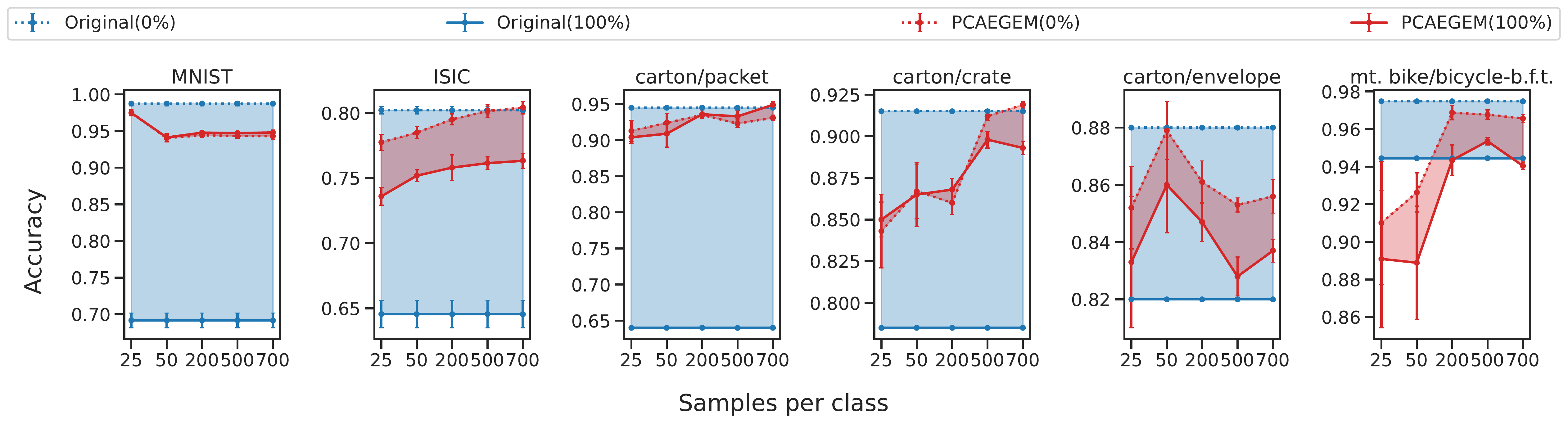}
    \caption{Effect of the number of instances used for refinement. The dotted line indicates clean-data accuracy whereas the solid line indicates 100\%-poisoned data accuracy. Mean and standard deviation are computed over 5 runs.}
    \label{fig:eval:samples}
\end{figure*}
As the number of instances available for refinement is limited, a natural question is what impact the number of samples has on the efficacy of refinement and if refining with too few instances can be detrimental. In this section, we repeat the experiment from Section~\ref{section:eval:accuracy} for datasets containing 25, 50, 200, 500, and 700 instances per class for refinement, under 0\% and 100\% uniform poisoning. Slack is again set to 5\%. If for some classes fewer correctly classified instances are available for refinement, these are over-sampled to achieve the desired number.

The effect of varying sample size is shown in Figure~\ref{fig:eval:samples}. It can be seen that especially in the low-sample regime, the positive effect of refinement is modulated by the number of instances. See~\app{app:samples} for all other methods.
While refinement with a small sample size appears in most cases to be remarkably effective for increasing 100\%-poisoned accuracy, clean-data accuracy tends to suffer as the sample does not cover all of the features necessary to generalize, some of which are thus pruned away.
For this reason, a larger refinement sample is in most cases beneficial, in particular for preserving clean-data accuracy. Yet, there are two cases that stand out as breaking this rule: 
The modified MNIST dataset and the `carton/envelope' task. In both cases, the gap between 0\% and 100\%-poisoned accuracy is close to constant, suggesting that the drop in accuracy stems from a loss of generalizing features rather than a loss of robustness, as could be induced e.g.\ by samples contaminated by CH features.
Considering the effect of slack, displayed in Fig.~\ref{fig:eval:slack}, we can see that those two scenarios are also cases for which 5\% slack leads to stronger than optimal refinement. We hypothesize that here, the negative effect of increasing sample size stems from the interrelation between sample size and refinement strength. In particular, for EGEM and PCA-EGEM, using fewer instances generally means less coverage of the feature space, which leads to fewer active neurons and hence more zero coefficients in the neuron-wise pruning factor described in Eq.~\ref{eq:activations-layerwise}. By this mechanism, larger sample sizes potentially lead to \textit{weaker} refinement. This effect may be exacerbated if a larger refinement sample also introduces images containing features that share parts of the latent representation with the CH feature.

Since clean-data accuracy can be evaluated on held-out data, we can observe that applying PCA-EGEM results in fairly predictable 100\%-poisoning performance across a wide range of sample sizes, i.e.\ the spread between clean-data and poisoned-data accuracy is small, as is demonstrated by the relatively small shaded area in Fig.~\ref{fig:eval:samples}.

A comparison to all other baselines can be found in Fig.~\app{app:fig:eval:samples}. What becomes particularly apparent is that Retrain benefits the most from large sample sizes and is among the weakest methods in the small-sample regime. On the modified MNIST dataset, its 100\%-poisoned accuracy only reaches 90\% with a refinement sample size of more than 200, whereas all other methods already achieve this with only 25 images.

\section{MNIST Revisited: Varying the Clever-Hans Feature}
\label{section:eval:mnist}
In this section, we are revisiting the MNIST task from Section~\ref{section:eval}, using different CH features. While all previous tasks contained localized additive artifacts, we will here investigate the effect of blur, removal of the lower part of the digits, and color shift as CH feature. We run the experiment with 5\% slack and 50 samples per class for refinement for each of these CH features.

\begin{figure*}[htb!]
    \centering
    \includegraphics[width=\textwidth]{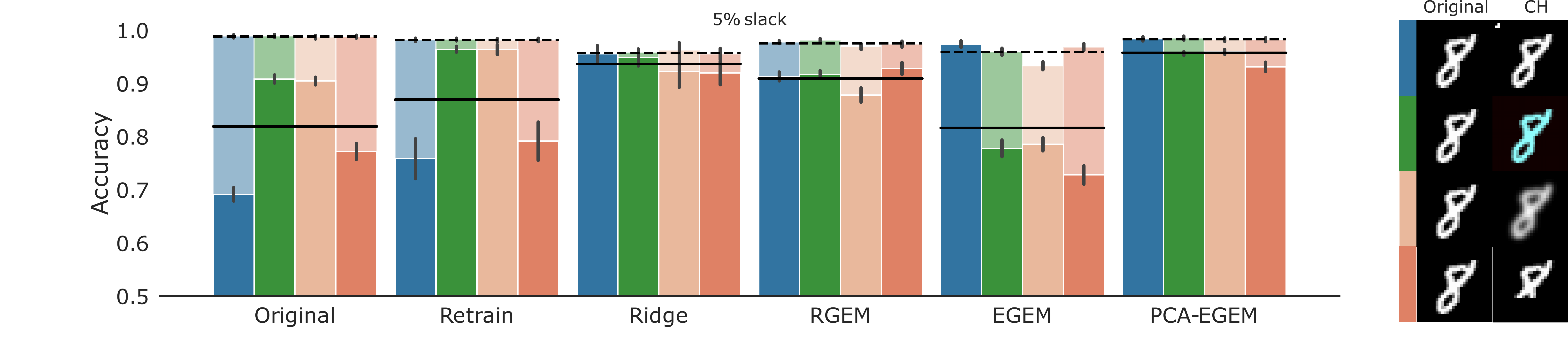}
    \caption{
    The accuracy for 0\% (lighter shade) and uniform 100\% (darker shade) poisoning with the spurious feature. Solid lines show average 100\%-poisoned accuracy and dashed lines show average clean-data accuracy over all datasets. The results shown are the mean accuracy and standard deviation obtained using 50 refinement samples per class on the respective test sets over 10 runs.
    }
    \label{fig:eval:mnist:pois_bars}
\end{figure*}

Fig~\ref{fig:eval:mnist:pois_bars} shows the classification accuracy under 0\% and 100\% poisoning and permits some interesting observations.

\paragraph{Spatially localized artifacts have a stronger effect} The original model sees the strongest decline in accuracy due to spatially localized CH features, specifically the additive artifact and removal. It appears that these features can be modeled well by the given architecture, e.g.\ the artifact can be captured easily by a convolutional filter or by the spatially disentangled features that the convolutional layers pass on to the fully-connected layers. This observation demonstrates that not all CH features are equal in their effect on a particular architecture, even if their rate of occurrence is equal.

\paragraph{EGEM underperforms on non-additive artifacts} The localized additive artifact is most amenable to removal via pruning whereas accuracy in all other scenarios drops when applying EGEM. We hypothesize that the pixel artifact has a very localized representation in the network (e.g.\ few particular convolutional filters, or neurons) and thus the neural basis, which EGEM uses for pruning, and the CH feature are more aligned. 

\begin{figure}[htb!]
    \centering
    \includegraphics[width=.4\textwidth]{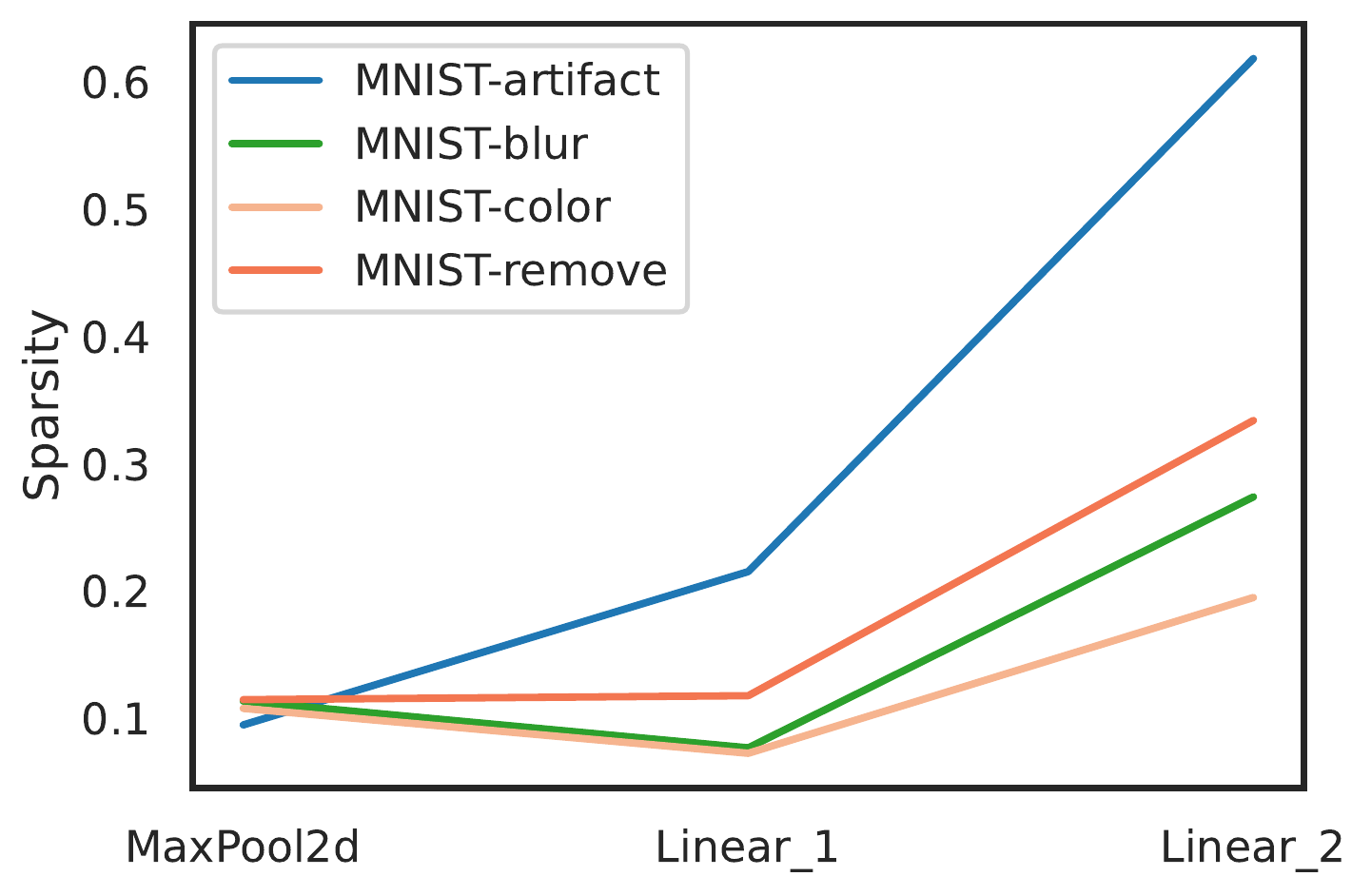}
    \caption{
    Sparsity of the change in representations of MNIST images of the class 8, induced by different CH artifacts and computed with Eq.\ \eqref{eq:sparsity}. Results are averaged over 100 images.
    }
    \label{fig:eval:mnist:sparse}
\end{figure}

In order to substantiate this hypothesis, we measure the sparsity of the change in representation induced by the CH feature. To this end, we sample 100 images of the digit 8 and measure the activations $a^l$ at every layer $l$ that is to be refined. We quantify the sparsity of the difference of activations before and after adding the CH artifact to the input by the ratio of the $\ell_2$ and the $\ell_1$ norm:
\begin{align}
    \frac{\|\mathbf{a}_{CH}^l - \mathbf{a}^l\|_2}{\|\mathbf{a}_{CH}^l - \mathbf{a}^l\|_1},
    \label{eq:sparsity}
\end{align}
which is 1 if the difference is localized at one dimension and less otherwise. The average sparsity of the effect of the CH feature is shown in Fig.~\ref{fig:eval:mnist:sparse}. The representation change induced by the spatially localized CH feature has the highest sparsity, which may explain why EGEM is effective in this scenario but not in others.

The representation realignment performed by PCA-EGEM via the PCA virtual layer resolves the problem, leading to consistently high accuracy across all types of CH features. This higher flexibility of PCA-EGEM w.r.t.\ the nature of the CH feature further justifies its use over basic EGEM.

\section{Use Case on CelebA: Reducing Bias}
\label{section:eval:celeba}
In this section, we will take a closer look at the effect of applying PCA-EGEM to a model trained on the CelebA dataset~\cite{LiuLWT2015}. In contrast to the previous experiments, we do not evaluate based on a specific known CH feature, but rather conduct the analysis in an exploratory manner, uncovering subpopulations for which a learned CH behavior is leading to biased classifications.
In practice, such an analysis could be done in hindsight, e.g.\ by proceeding along the following steps: 1) a third-party model is acquired, 2) it is refined using PCA-EGEM and available data, 3) it is deployed, i.e.\ applied to test data, 4) the effect of PCA-EGEM is examined on the test data using XAI and recall statistics.

The CelebA dataset contains 202,599 portrait images of celebrities, each associated with 40 binary attributes. The existence of spurious correlations in the CelebA dataset has been documented previously~\cite{SagawaRKP2020, XuWKGBF2020, Khani2021,SagawaKHL2020} and it can be seen in~\app{app:data:celeba:corr} that the attributes in the training set are correlated to various degrees. We train a convolutional neural network (details in~\app{app:train}) on the `train' split of the CelebA dataset using cross-entropy loss on a `Blond\_Hair'-vs-not classification task. The training data is stratified and we achieve a binary test accuracy of 93\%, which is comparable to accuracy reported in other works, e.g.\ Sagawa et al.~\cite{SagawaKHL2020}. We regard this classifier as a model given to the user by a third party.

In the following, we will assume a scenario where the user seeks to use the third-party classifier to retrieve blond people from a set of images available during deployment. They wish this retrieval process to be accurate and not biased against subgroups in the population. In order to analyze the impact of applying PCA-EGEM on such retrieval task, we simulate a validation set where the user has a limited subset of `clean' examples, specifically, 200 examples of both classes, that are correctly predicted by the model and whose explanation highlight the actual blond hair as determined by LRP scores falling dominantly within the area of the image where the hair is located (see~\app{app:experiments}). These explanations (considered by the user to be all valid) are then fed to PCA-EGEM to produce a model that is more robust to potential unobserved Clever Hans effects. As for the previous experiments, we use 5\% slack, which translates here to $\alpha=0.01$. 

\subsection{PCA-EGEM Reduces Exposure to Shirt Collars}
\label{section:eval:celeba:collar}
\begin{figure}[t]
    \includegraphics[width=\linewidth]{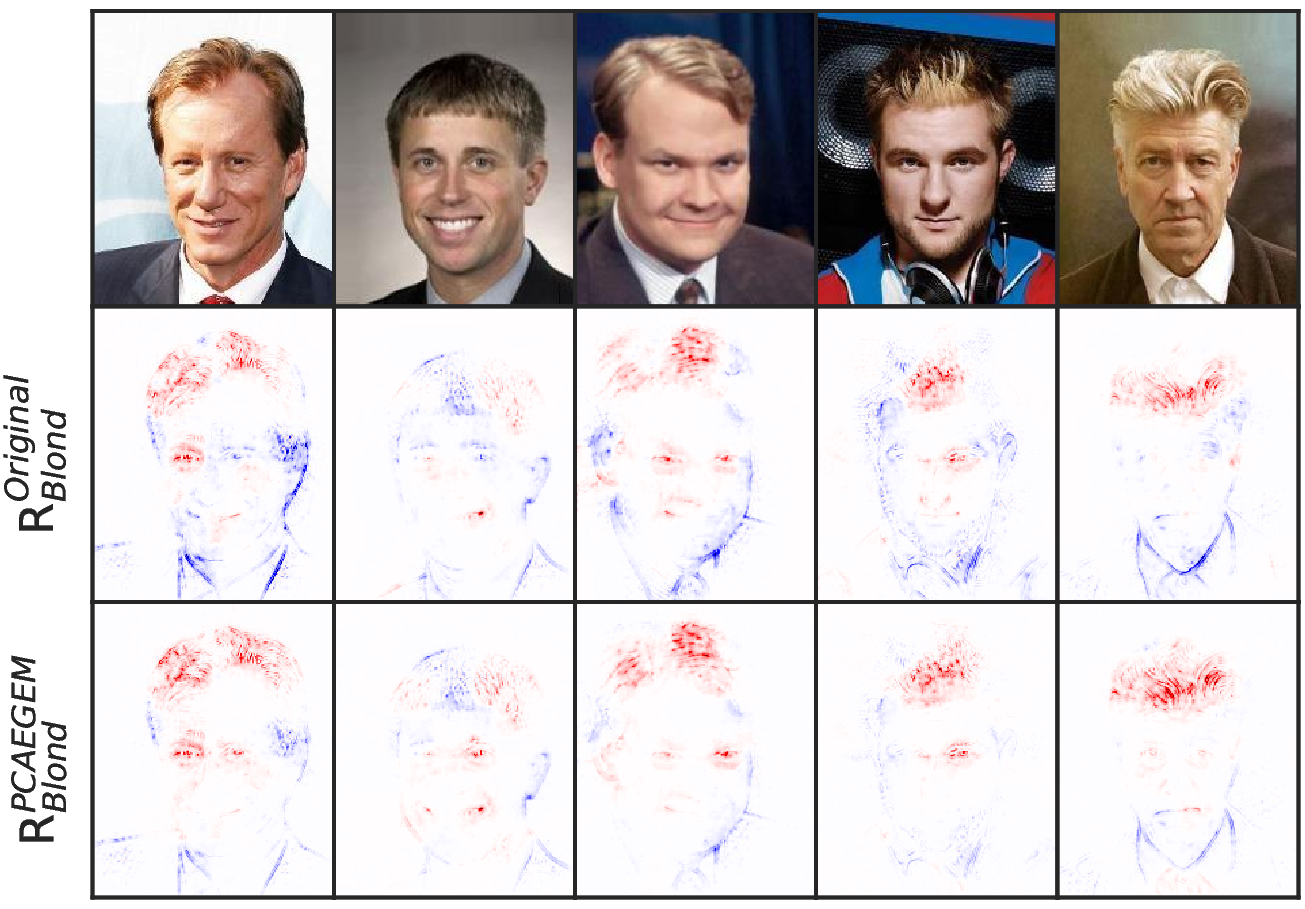}
\caption{Test set images that exhibit strong changes in the detection of blond hair and corresponding LRP explanations, before and after refinement. Red indicates positive and blue negative contribution to the detection of blond hair. Shirt collars and similar features appear to inhibit the prediction of blond hair in the original model but less so in the refined one.} \label{fig:celeba:results}
\end{figure}
After the model is deployed, the analysis of the decision strategy (of the original and refined model) can be reexamined in light of the new data now available. Fig.~\ref{fig:celeba:results} shows explanations for some retrieved images, specifically, evidence for them being predicted to be blond. We can observe that pixels displaying hair are considered to be relevant and remain so after refinement.

However, one can also identify a significant change of strategy before and after refinement in the lower part of the image: The original model appears to make heavy use of shirt and suit collars as a feature inhibiting the detection of blond hair, whereas such inhibiting effect is much milder in the refined model. This observation suggests that PCA-EGEM has effectively mitigated a previously unobserved Clever Hans strategy present in the original model, and as a result, effectively aided the retrieval of images with collars on them.

\subsection{PCA-EGEM Balances Recall Across Subgroups}
\begin{figure*}[t]
    \centering
    \includegraphics[width=.8\linewidth]{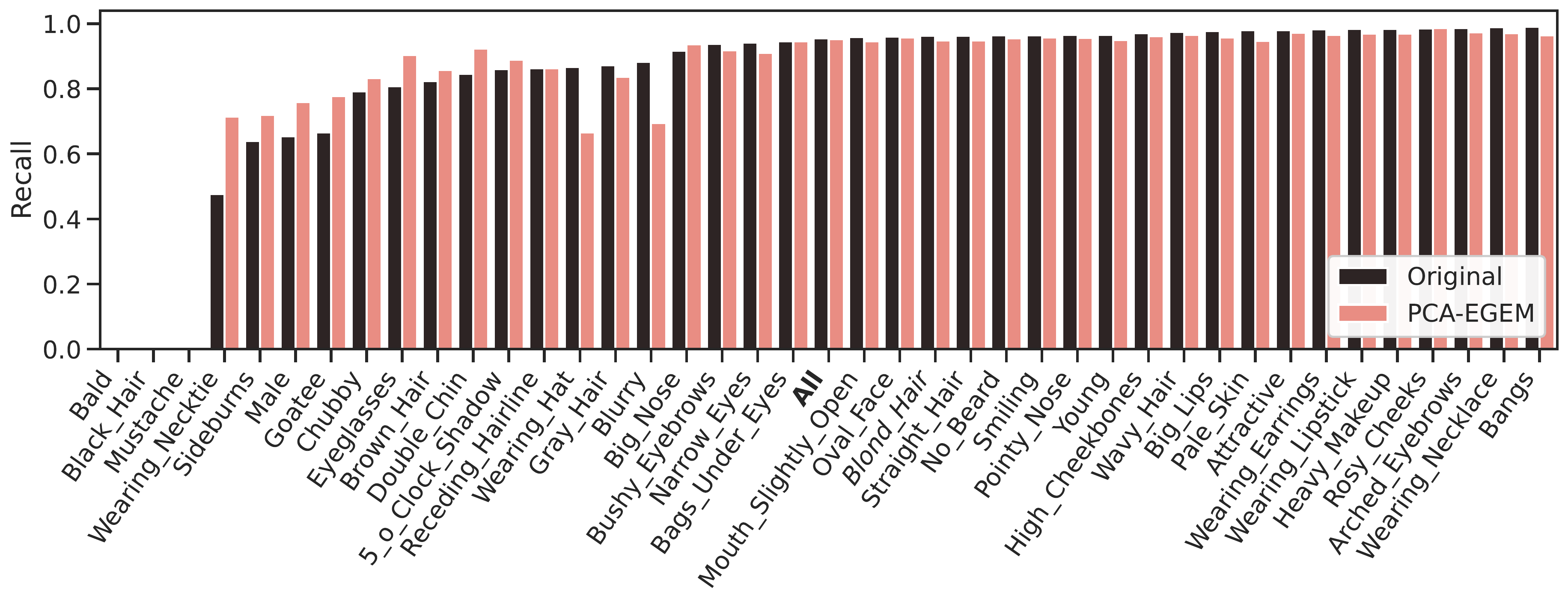}
\caption{Comparison of recall of the pretrained model before (Original) and after refinement (PCA-EGEM). Recall is calculated on subsets of CelebA containing only samples exhibiting the attribute on the x-axis. `All' is sampled from the whole test set.} \label{fig:celeba:recall}
\end{figure*}
We will now analyze the implication of the Clever Hans effect reduction by PCA-EGEM on specific subgroups, specifically, whether certain subgroups benefit from the model refinement in terms of recalling members with the attribute `Blond\_Hair'.

To this end, we randomly sample for every attribute in the dataset, a subset of 5000 images from the test data that only contains samples exhibiting this attribute. If fewer images are available for some attribute, we use all of them. We evaluate the classifier before and after the application of PCA-EGEM on each of these subgroups.

Figure \ref{fig:celeba:recall} shows recall scores for each subgroup.
We observe a substantial increase of recall on low-recall subgroups, such as `Wearing\_Necktie', `Goatee', and `Male'. Most high-recall groups see only minuscule negative effects. Overall, while having almost no effect on the dataset-wide recall, we can observe that the application of PCA-EGEM rebalances recall in favor of under-recalled subsets. Our investigation thus demonstrates that a model bias responsible for under-detecting blond hair in these subgroups has been mitigated by applying PCA-EGEM. This consequently leads to a set of retrieved images that is more representative of the different subgroups and more diverse.

It is of theoretical interest to ask whether such rebalancing effect would generalize to other scenarios. An argument is that the underrepresentation of certain subgroups in the retrieved set is mainly caused by subgroups with low prevalence of the class of interest and those subgroups being actively suppressed by the model in order to optimize its accuracy. In practice, such suppression can be achieved by identifying features specific to the subgroup and, although causally unrelated to the task, making these features contribute negatively to the output score. Our PCA-EGEM technique, by removing such task-irrelevant Clever Hans features, remodels the decision function in favor of these low-prevalence subgroups, thereby leading to a more balanced set of retrieved instances.

Two outliers to the overall rebalancing effect can however be noted in Fig.\ref{fig:celeba:recall}: `Wearing\_Hat', and `Blurry'. Interestingly, these are two subgroups in which the feature of interest (the hair) is occluded or made less visible. In other words, in these two subgroups, only weakly correlated features are available for detection, and their removal by PCA-EGEM consequently reduces the recall. An underlying assumption behind the rebalancing effect is therefore that the concept of interest (blond hair) is detectable in the input image without resorting to weakly or spuriously correlated features which may be refined away due to being underrepresented in the available data. A more detailed discussion can be found in~\app{app:exp:celeba}.

Overall, we have demonstrated in our CelebA use case, that PCA-EGEM can be useful beyond raising accuracy on disadvantageous test-set distributions. Specifically, we have shown that our PCA-EGEM approach enables the retrieval of a more diverse set of positive instances from a large heterogeneous dataset.

\section{Use Case: Removing a CH Effect in Movie Reviews}
\label{section:eval:movies}
In this section, we demonstrate the use of PCA-EGEM for unlearning CH features in the context of binary sentiment classification of movie reviews, employing a transformer model.

Existing works have underscored the susceptibility of pretrained sentiment classifiers to features spuriously correlated with the class, such as actor names~\citep{Ali2022, Schnake2022} or the occurrence of stop words~\citep{Liusie2022}, even if these words should not be taken into account for the classification process. 

To demonstrate the potential of PCA-EGEM to reduce such biases, we fine-tune a pretrained DistillBert~\citep{Sanh2019} model~\footnote{\href{https://huggingface.co/distilbert-base-uncased-finetuned-sst-2-english}{https://huggingface.co/distilbert-base-uncased-finetuned-sst-2-english}} on the training set of the IMDB dataset~\citep{Maas2011}. A CH effect is artificially induced by appending the sentence ``\textit{But that's just my unrefined opinion.}'' to 25\% of positive reviews within the training set. This sentence was chosen as it is plausible in the context of movie reviews and may be a formulation linked to specific reviewers rather than sentiment, thereby potentially giving rise to a CH effect.
For refinement with PCA-EGEM we use 1000 positively and 1000 negatively labeled reviews, randomly selected from the validation set. Employing a slack of 5\%, we refine the linear output layer of each multi-head self-attention block, using the same hyper-parameter search procedure described in Section~\ref{sec:eval:hyper}.

The unrefined fine-tuned classifier achieves a test accuracy of 92.4\% whereas the refined version achieves 90.1\%. Upon appending the CH sentence to all test-samples (100\% uniform poisoning), the unrefined model only achieves the base-rate accuracy of 50\% whereas the refined model maintains high accuracy of 88.8\%. 
\begin{figure}[t!]
    \centering
    \includegraphics[width=\linewidth]{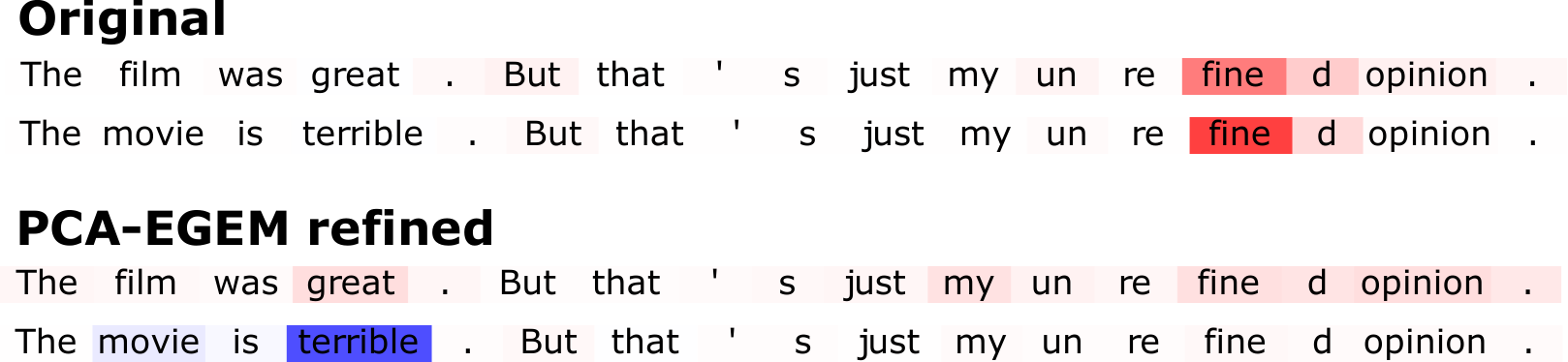}
\caption{
Effect of each word, estimated by replacing them by the \texttt{UNK} token. Red: positive, blue: negative sentiment. Normalized on the sentence level.
} \label{fig:imdb}
\end{figure}
Evidently, the refinement procedure was able to mitigate sensitivity to the introduced CH feature. We create explanations for two synthetic, poisoned review examples in Fig~\ref{fig:imdb} by measuring the effect of replacing each token in the sentence with the \texttt{UNK} token. The original model focuses primarily on the appended sentence, specifically, the token ``fine'' which DistillBert seems to encode saliently for its relevance to sentiment classification\footnote{When repeating the same experiment without the token ``fine'' (i.e.\ by appending the shorter statement ``\textit{But that's just my opinion.}''), the CH effect does not develop.}. On the other hand, this effect is starkly reduced in the refined model.
This experiment demonstrates that the proposed PCA-EGEM refinement approach can also be applied insightfully to other neural network architectures, such as transformer models, and to data modalities beyond images, such as text.

\section{Open Questions}
We could demonstrate the efficacy of our proposed EGEM and PCA-EGEM refinement methods for mitigation of Clever Hans effects in Sections~\ref{section:eval}--\ref{section:eval:movies}, however, it could also be observed that 1) a complete removal of the model's response to the spurious (CH) feature is usually not fully achieved, and 2) classification accuracy on clean data may suffer slightly. We suggest that there are multiple reasons for these undesired effects. 

\subsection{Entangled Feature Representations}
In deep neural networks, CH features are not generally neatly disentangled from generalizing features.
This means that either entangled well-generalizing features might suffer from pruning, reducing clean-data accuracy, or CH features might not be pruned due to being entangled with a feature present in the clean dataset. The latter would inhibit robustification against the CH feature. In this case, a trade-off between clean-data accuracy and robustness exists, which can be navigated via the slack parameter introduced in Section~\ref{sec:eval:hyper}. 

A way to avoid this conflict altogether could be a feature disentanglement strategy, separating the CH feature from others. While the PCA-EGEM extension achieves a basic form of disentanglement, more refined disentanglement methods based, for example, on finding independent components, could be considered in future work. Chormai et al.~\citep{Chormai2022} have demonstrated the efficacy of explanation-based disentanglement into independent subspaces. In principle, our proposed soft-pruning strategy could be applied in such spaces. An alternative direction may be the disentanglement into factors derived from human behavioral data~\citep{Muttenthaler2022}, which have been shown to be represented in different pretrained neural networks to various degrees~\citep{Muttenthaler2023}.

\subsection{Limited Available Data}
Another open question is how the proposed exposure minimization approach can be effective in the case where the available data is scarce or unrepresentative of the true input distribution.

The disentanglement methods discussed above, beyond achieving the desired separation between the CH and correct strategies, may also be useful for addressing data scarcity. In particular, they could enable the generation of low-dimensional subspaces in which only a few data points are sufficient to characterize the correct decision strategy.

Furthermore, methods that extend the data in an informed manner could also be useful. This includes methods for synthesizing class prototypes (via activation-maximization ~\citep{hansen2011visual, Nguyen2019}, or class-conditioned sampling~\citep{DBLP:journals/corr/NguyenYBDC16}). The generated prototypes are sometimes referred to as `\textit{global explanation}'. These generated instances may be appended to the available data and subsequently included in the dataset available for refinement.

\subsection{Imprecise Explanations or User Assessment}
Beyond the need for sufficient data to validate the model, it is also crucial that the XAI method upon which our exposure minimization technique builds is able to accurately convey the presence or absence of CH effects.
While there have been concerns about the effectiveness of XAI methods in some scenarios~\cite{Adebayo2018, Dombrowski2019, Adebayo2022, Binder_2023_CVPR}, especially their ability to reliably detect CH effects, these concerns could be addressed in future work by moving beyond pixel-wise explanations e.g.\ using concept-based or counterfactual explanations~\cite{Stepin2021, Verma2021, Dombrowski2022}. 

Furthermore, assuming a well-working XAI technique, the user should be able to make a decision (e.g.\ no CH effects) in an error-free manner. Lack of attention, limited knowledge on how to interpret explanations, or an inability to process the potentially large amount of produced explanations in reasonable time may cause the user to falsely conclude that the ML model at hand is CH-free. One overall direction to address these issues is to develop improved interaction tools between the XAI technique and the user. For example, the set of explanations could be rendered to the user in a more intuitive way so that more examples can be inspected and outliers can be paid special attention -- a strategy already implemented e.g.\ in SpRAy~\citep{Lapuschkin2019}. Beyond that, the amount of explanations to be inspected by the user could be decreased by extracting a representative subset of instances from the available data. This may be done by summarizing techniques such as coresets~\citep{phillips2017coresets} or pool-based sampling approaches as are used in active learning~\citep{Ren2021} and may also be based on explanations~\citep{Mahapatra2021}.

\subsection{Performance After Removing CH Features}
As pointed out in previous work~\citep{Khani2021}, removing spuriously correlated (CH) features can hurt performance on data where this correlation holds.
One such effect was observed in our experiment on CelebA in Section~\ref{section:eval:celeba}, where a refined model would underperform on the `Blurry' and `Wearing\_Hat' subgroups at test time due to not being able to rely on the weakly correlated CH features. If the CH features were present in the data available for refinement, a proper reweighting of features (inhibiting CH features and enhancing correct ones) would in principle be achievable before deployment (e.g.\  \cite{anders2022finding, Kirichenko2022}). In the more difficult scenario studied in this paper, this is not possible. Still, one could potentially monitor the discrepancy between the original model and PCA-EGEM at test time, in order to quickly mitigate unresolved biases in the deployed model.

\section{Conclusion}
Sensitivity to distribution shifts, such as the ones induced by spurious correlations (so-called Clever Hans effects) has long been an Achilles heel of machine learning approaches, such as deep learning. The problem becomes even more pronounced with the increasing adoption of foundation models for which the training data may not be public and is thus closed to scrutiny.
Explanation techniques have the potential to uncover such deficiencies by putting a human in the loop~\cite{Lipton2016, Lapuschkin2019, Murdoch2019, Samek2019a}. Previous work in XAI has mainly focused on improving explanations or fixing flaws in a model that have been identified by the user from such explanations. In contrast, we have considered the under-explored case where the human and the explanation \textit{agree} but where there are possibly unobserved spurious features that the model is sensitive to. While recent work has shown that XAI-based validation techniques may fail to detect some of these Clever Hans strategies employed by a model~\cite{Adebayo2022}, we have argued that one can nevertheless still reduce the exposure of a model to some of these hidden strategies and demonstrated this via our proposed Explanation-Guided Exposure Minimization approach. 

Our approach, while formulated as an optimization problem, reduces to simple pruning rules applied in intermediate layers, thereby making our method easily applicable, without retraining, to complex deep neural network models such as those used in computer vision. Our method was capable of systematically improving prediction performance on a variety of complex classification problems, outperforming existing and contributed baselines.

Concluding this paper, we would like to emphasize the novelty of our approach, which constitutes an early attempt to leverage correct explanations for producing refined ML models and attempts to tackle the realistic scenario where Clever Hans features are not accessible. We believe that in future work, the utility derived from explanations via refinement can still be expanded, e.g.\ by letting the user specify what is correct and what is incorrect in an explanation so that the two components can be treated separately, or by identifying sets of examples to present to the user that are the most useful to achieve model refinement, for example, by ensuring that they cover the feature space adequately or by active learning schemes.

\section{Acknowledgements}
This work was partly funded by the German Ministry for Education and Research (under refs 01IS14013A-E, 01GQ1115, 01GQ0850, 01IS18056A, 01IS18025A and 01IS18037A), the German Research Foundation (DFG) as Math+: Berlin Mathematics Research Center (EXC 2046/1, project-ID: 390685689). Furthermore, KRM was partly supported by the Institute of Information \& Communications Technology Planning \& Evaluation (IITP) grants funded by the Korean government (MSIT) (No. 2019-0-00079, Artificial Intelligence Graduate School Program, Korea University and No. 2022-0-00984, Development of Artificial Intelligence Technology for Personalized Plug-and-Play Explanation and Verification of Explanation). We thank Pattarawat Chormai and Christopher Anders for the valuable discussion and the extracted watermark artifact.

\putbib
\end{bibunit}

\newpage

\input{supplement}

\end{document}

%% file: supplement.tex

\resetTitleCounters
\makeatletter

\long\def\MaketitleBox{%
  \resetTitleCounters
  \def\baselinestretch{1}%
  \begin{center}%
   \def\baselinestretch{1}%
    \Large{Preemptively Pruning Clever-Hans Strategies in Deep Neural Networks\\[2mm]\large \textsc{(Supplementary Notes)}}\par\vskip18pt
    \normalsize{Lorenz Linhardt, Klaus-Robert M\"uller, Gr\'egoire Montavon}\par\vskip10pt
    \footnotesize\itshape
    \par\vskip36pt
    \end{center}%
  }
\makeatother
\allowdisplaybreaks

\makeatletter
\def\ps@pprintTitle{%
  \let\@oddhead\@empty
  \let\@evenhead\@empty
  \def\@oddfoot{\reset@font\hfil\thepage\hfil}
  \let\@evenfoot\@oddfoot
}
\makeatother

\sloppy

\appendix
\onecolumn 
\MaketitleBox

\setcounter{page}{1}

\begin{bibunit}

\section{Structure of Explanations}
\label{app:relevance-structure}
In this section, we want to show that, assuming a common neural network with neurons of the type
\begin{align}
z_j &= \textstyle \sum_i a_i w_{ij} + b_j \nonumber\\
a_j &= g(z_j),
\end{align}
attribution scores associated with the explanation techniques \gi~\citep{Shrikumar2017}, 
Integrated Gradients~\cite{Sundararajan2017}, and layer-wise relevance propagation (LRP)~\cite{Bach2015,Montavon2019a} can be decomposed and written in the form:
\begin{align}
    R_i &= \textstyle \sum_j R_{ij}\nonumber\\
    R_{ij} &= a_i \rho(w_{ij}) d_j \label{eq:structure}
\end{align}
where $\rho:\mathbb{R} \to \mathbb{R}$ is some function, and $d_j$ is a term that only indirectly depends on the activation $a_i$ and the weight $w_{ij}$.

\subsection{\texorpdfstring{\gi}{Gradient-Times-Input}}
\label{sec:struct-gi}
Denoting by $y$ the output of the neural network that we would like to explain, we can write the scores obtained by \gi{} w.r.t.\ any layer with activations $\{a_i\}_{i=1}^N$ as:
\begin{align}
    R_i &= a_i \pdv{y}{a_i}.
\intertext{Using the chain rule, the equation can be further developed as:}
    &= a_i \sum_j \pdv{y}{z_j} \pdv{z_j}{a_i} = \sum_j \underbrace{a_i \underbrace{\pdv{z_j}{a_i}}_{w_{ij}} \underbrace{\pdv{y}{z_j}}_{d_j}}_{R_{ij}} 
\end{align}   
form which we can identify the desired structure of Eq.\ \eqref{eq:structure}.

\subsection{Integrated Gradients}
\label{sec:struct-ig}
For the Integrated Gradients formulation, we consider as integration path a segment from the origin to the data point (i.e.\ the map $t \mapsto t \cdot \ba$ with $t \in [0,1]$):
    \begin{align}
        R_i &= \int \pdv{y}{a_i}\pdv{a_i}{t} dt
\intertext{Using the chain rule, the equation can be further developed as:}
        &= \int \sum_j \pdv{y}{z_j}\pdv{z_j}{a_i}\pdv{a_i}{t} dt\\
        &= \sum_j \int \pdv{y}{z_j}\pdv{z_j}{a_i}\pdv{a_i}{t} dt\\
        &= \sum_j \underbrace{\underbrace{\Big(\int \pdv{y}{z_j} dt\Big)}_{d_j} w_{ij} a_i}_{R_{ij}} 
    \end{align} 
From which we can again identify the structure of Eq.\ \eqref{eq:structure}.

\subsection{Layer-wise Relevance Propagation}
\label{sec:struct-lrp}
Starting from a generic LRP rule that admits standard LRP rules such as LRP-0, LRP-$\gamma$, and LRP-$\epsilon$ as special cases, specifically:
\begin{align}
    R_i &= \sum_j \frac{a_i\rho(w_{ij})}{\sum_{i'}a_{i'}\rho(w_{i'j}) + \epsilon} \cdot R_j \\
\intertext{with $\rho(t) = t + \gamma \max(0,t)$ with $\gamma$ nonnegative, we get after some slight reordering of the equation}
    &= \sum_j \underbrace{a_i \rho(w_{ij}) \underbrace{\frac{R_j}{\sum_{i'}a_{i'}\rho(w_{i'j}) + \epsilon}}_{d_j}}_{R_{ij}}
\end{align}
which has the structure of Eq.\ \eqref{eq:structure}.

\section{Derivation of EGEM}
\label{app:egem:derivation}
In this section, we derive a closed-form solution for the EGEM method, which we stated in Section~3 of the main paper as
\begin{align}
\forall_{ij}:~w_{ij} = \frac{\mathbb{E}[a_i^2d_j^2]}{\mathbb{E}[a_i^2d_j^2] + \lambda \mathbb{E}[d_j^2]} w_{ij}^\text{old}
\end{align}
with the purpose of solving the objective
\begin{align}
\min_w ~ \sum_{ij} \mathbb{E}  \big[ (R_{ij}(w_{ij}) - R_{ij}(w_{ij}^\text{old}) )^2 + \lambda (s_{ij}(w_{ij}))^2 \big]
\end{align}
where
\begin{align}
R_{ij}(w_{ij}) &= a_i \, \rho(w_{ij}) \, d_j
\intertext{and}
s_{ij}(w_{ij}) &= \rho(w_{ij}) \, d_j.
\end{align}
Substituting these last two terms into the objective, we get:
\begin{align}
\min_w ~ \sum_{ij} \mathbb{E}\big[\big(a_i \, \rho(w_{ij}) \, d_j - a_i \, \rho(w_{ij}^\text{old}) \, d_j\big)^2 + \lambda \cdot \big(\rho(w_{ij}) \, d_j\big)^2\big]
\end{align}
We observe that each term of the sum depends on its own parameter $w_{ij}$. Hence, each term can be minimized separately. Consider one such term and compute its gradient:
\begin{align}
E_{ij}(w_{ij}) &= \mathbb{E}\big[\big(a_i \, \rho(w_{ij}) \, d_j - a_i \, \rho(w_{ij}^\text{old}) \, d_j\big)^2 + \lambda \cdot (\rho(w_{ij}) \, d_j)^2 \big] & \\
\nabla E_{ij}(w_{ij}) &= \mathbb{E}\big[ 2 \big(a_i \, \rho(w_{ij}) \, d_j - a_i \, \rho(w_{ij}^\text{old}) \, d_j\big) \cdot a_i \, \rho'(w_{ij})\, d_j + 2\lambda \cdot \rho(w_{ij}) \, \rho'(w_{ij}) \, d_j^2\big]
\end{align}
We now find where the gradient is zero. Our derivation uses the fact that $\rho(w_{ij})$ and its derivative do not depend on the data and can therefore be taken out of the expectation:
\begin{align}
\mathbb{E}\big[ 2 \big(a_i \, \rho(w_{ij}) \, d_j - a_i \, \rho(w_{ij}^\text{old}) \, d_j\big) \cdot a_i \, \rho'(w_{ij})\, d_j + 2\lambda \cdot \rho(w_{ij}) \, \rho'(w_{ij}) \, d_j^2\big]  &\overset{!}{=} 0 \\
\mathbb{E}\big[ \big(a_i \, \rho(w_{ij}) \, d_j - a_i \, \rho(w_{ij}^\text{old}) \, d_j\big) \cdot a_i \, d_j + \lambda \cdot \rho(w_{ij}) \, d_j^2\big]  &\overset{!}{=} 0 \\
\rho(w_{ij}) \, \mathbb{E}[a_i^2 d_j^2] - \rho(w_{ij}^\text{old}) \, \mathbb{E}[a_i^2 d_j^2] + \lambda \rho(w_{ij}) \, \mathbb{E}[d_j^2]  &\overset{!}{=} 0 \\
\rho(w_{ij}) \, \mathbb{E}[a_i^2 d_j^2] + \lambda \cdot \rho(w_{ij}) \mathbb{E}[d_j^2]  &\overset{!}{=} \rho(w_{ij}^\text{old}) \, \mathbb{E}[a_i^2 d_j^2] \\
\rho(w_{ij}) &\overset{!}{=} \frac{\mathbb{E}[a_i^2  d_j^2]}{ \mathbb{E}[a_i^2 d_j^2] + \lambda \cdot \mathbb{E}[ d_j^2]} \cdot \rho(w_{ij}^\text{old})
\end{align}
Furthermore, the equation above also implies
\begin{align}
w_{ij} &\overset{!}{=} \frac{\mathbb{E}[a_i^2  d_j^2]}{ \mathbb{E}[a_i^2 d_j^2] + \lambda \cdot \mathbb{E}[ d_j^2]} \cdot w_{ij}^\text{old}
\end{align}
for the choices of the function $\rho$ encountered in~\app{app:relevance-structure}.

\section{Data}
\label{app:data}
\subsection{Modified MNIST}
The modified MNIST is a variant of the original MNIST dataset~\cite{mnist} where a small 3-pixel artifact is pasted onto the top left corner of 70\% of the images of the digit `8'. In order to generate a somewhat natural split of the data where one part is free of the artifact, we train a variational autoencoder~\cite{Kingma2014} on the artifact-modified dataset to model the underlying distribution. Then, we manually select a single dimension and threshold along this dimension to separate the data into two partitions, such that both partitions contain all digits, but the modified samples only fall onto one side. This partitioning defines the pool of clean samples from which to draw the refinement data. 

\subsection{Correlation Structure in CelebA}
\label{app:data:celeba:corr}
\begin{figure}[htb!]
    \centering
    \includegraphics[width=.8\linewidth]{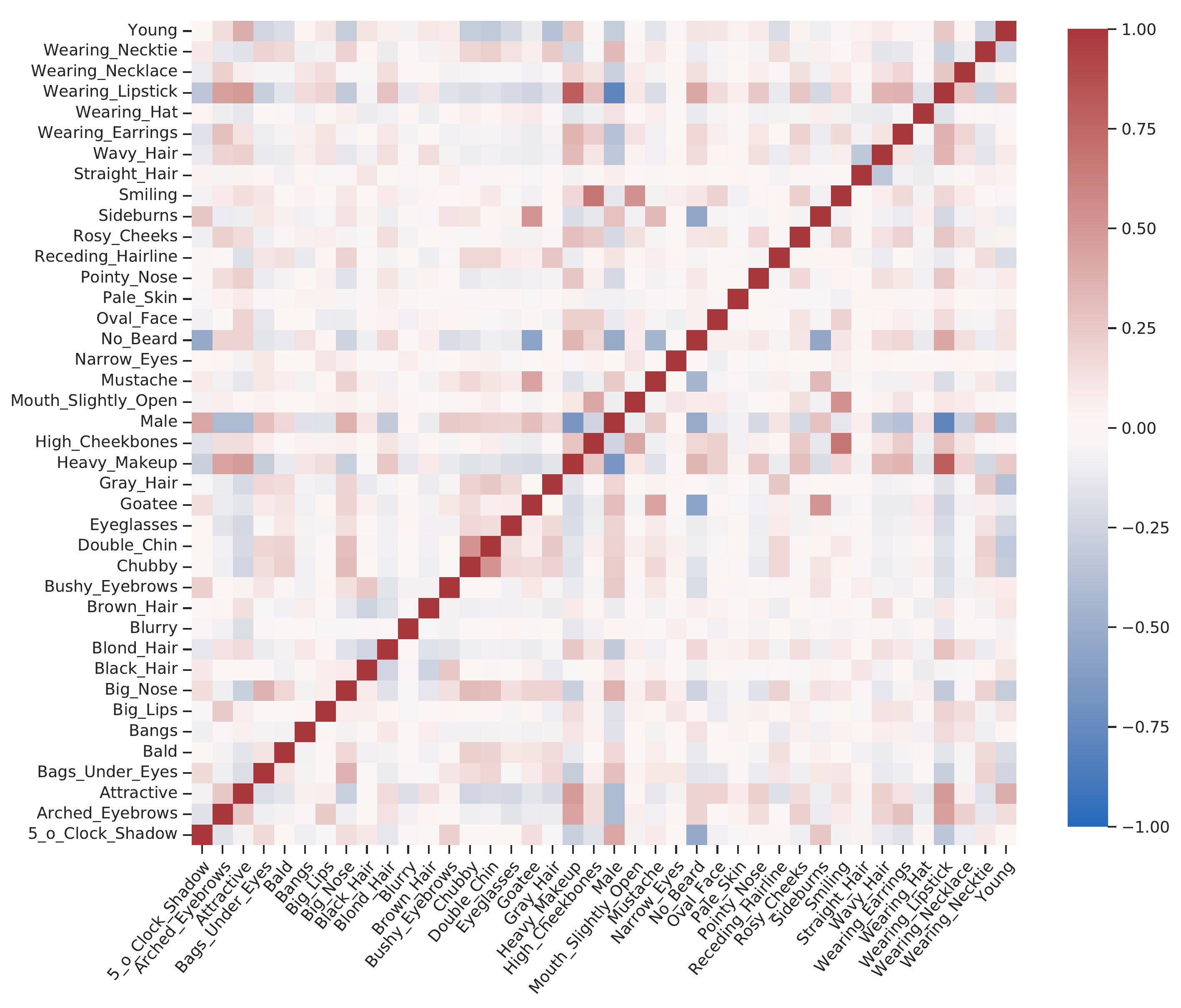}
    \caption{Correlation matrix of attributes in the CelebA training data.}
    \label{app:fig:data:celeba:correlation}
\end{figure}

In this section, we present the correlation structure of the attributes in the training partition of the CelebA dataset~\cite{LiuLWT2015}, which is essential when reasoning about Clever Hans effects since they are based on spurious correlations in the training data. Fig.~\ref{app:fig:data:celeba:correlation} displays the respective correlation matrix.
Notably, `Male' and `Blond\_Hair' are negatively correlated, while `Male' and `Wearing\_Necktie', `Goatee', and `Sideburns' are positively correlated, which could lead `Male'-related visual features to be used as an inhibitory signal by models trained to detect blond hair on this dataset.

\section{Response-Guided Exposure Minimization (RGEM)}
\label{app:rgem}
Consider the problem of learning a low-complexity model that reproduces the output of some original model on the validated data. Let $f(\x, \w)$ and $f(\x, \w_\text{old})$ be the output of the student and teacher models respectively, parameterized by weights $\w$ and $\w_\text{old}$. We can formulate the objective as the optimization problem:
\begin{align}
    \min_{\w} ~~ \mathop{\mathbb{E}}[(f(\x, \w) - f(\x, \w_\text{old}))^2] + \lambda \|\w\|^2
\label{app:eq:rgem}
\end{align}
where the expectation is computed over the available data. The first term ensures the reproduction of the model output on the refinement data. The second term with regularization parameter $\lambda$ penalizes overall model exposure, e.g.\ forcing the model to not be too complex. For the linear case, where the teacher model and student model are given by $f(\x, \w_\text{old}) = {\w_\text{old}}^\top \x$ and $f(\x, \w) = \w^\top \x$ respectively, we get a closed form solution by equating the objective's gradient to zero:
\begin{align}
\w = (\Sigma + \lambda I)^{-1} \Sigma \w_\text{old}
\label{app:eq:weights-output}
\end{align}
where $\Sigma = \mathbb{E}[\x\x^\top]$. This equation resembles the ridge regression solution; the difference being that the cross-covariance between data and targets $\mathbb{E}[\x \y]$ in the original model is replaced by the term $\Sigma \w_\text{old}$. This term performs a realignment of the pretrained model's weights along the validation data. This realignment with the refinement data, desensitizes the model to directions in feature space that are not expressed in the available data and that the user could not verify, giving some level of immunity against a possible CH effect in the classifier. For neural networks with a final linear projection layer, response-guided exposure minimization (RGEM) can be applied to this last layer.

We can rewrite Eq.~\eqref{app:eq:weights-output} as ridge regression on the predictions $f(\x, \w_\text{old})$:
\begin{align}
\w = (\X^T\X + \lambda I)^{-1} \X^T f(\X, \w_\text{old})
\end{align}
where $f(\X, \w_\text{old})$ is the vector of outputs of the original model on the refinement data. 

\section{Models: Training Procedure}
\label{app:train}
In this section, we provide the details of the training procedure relevant for reproducing the models used in our experiments in Section~4 of the main paper.

For the experiments on ImageNet and the ISIC dataset, we use the pretrained VGG-16 network, as provided in the pytorch library~\cite{Paszke2019}. For the ISIC dataset, only keep the first two output nodes and fine-tune the network for 10 epochs with learning rate 0.0001 and batch size 64. For the experiments on MNIST, we train a neural network specified in Table~\ref{app:tab:net mnist} for 5 epochs with learning rate 0.001 and batch size 128. For the experiments on the CelebA dataset, we train a neural network specified in Table~\ref{app:tab:net celeba} for 10 epochs with a learning rate of 0.000025 and a batch size of 100. 
For all training procedures, the Adam optimizer~\cite{Kingma2015Adam} is used.
\begin{table}[!htb]
    \centering
    \begin{tabular}{ |c|l|c|c|c| }
     \hline
     ID & Type & Channels & Kernel & Stride \\
     \hline
    1 & Conv. (ReLU) & 8 & 3 & 1 \\
    2 & Max-pool & - & 2 & - \\
    3 & Conv. (ReLU) & 16 & 5 & 1 \\
    4 & Max-pool & - & 2 & - \\    
    5 & FC (ReLU) & 200 & - & - \\
    6 & FC (Identity) & 10 & - & - \\
     \hline
    \end{tabular}
    \caption{Neural network architecture used for the experiments using the MNIST dataset.}
    \label{app:tab:net mnist}
\end{table}
\begin{table}[!htb]
    \centering
    \begin{tabular}{ |c|l|c|c|c| }
     \hline
     ID & Type & Channels & Kernel & Stride \\
     \hline
    1-11 & 3 VGG-16 blocks & - & - & - \\
    12 & Conv. (ReLU) & 128 & 3 & 1 \\
    13 & Adaptive max-pool & - & - & - \\
    14 & FC (ReLU) & 512 & - & - \\
    15 & FC (Identity) & 2 & - & - \\
     \hline
    \end{tabular}
    \caption{Neural network architecture used for the experiments using the ISIC dataset.}
    \label{app:tab:net celeba}
\end{table}

\section{User Verification and Experimental Details}
\label{app:experiments}
In the following we provide details of the simulated user verification procedure and of the generation of the poisoned data sets, where the relation between the class and the spurious features is modified.

\subsection{Simulation of User-Verification}
Generally, we would consider samples to be user-verified if they are correctly predicted by the model and if a human has inspected and agrees with the corresponding explanation.

For the experiments in Section~4 of the main paper we assumed that all correctly classified clean samples are user-verified and thus just removed samples containing the artifact from the refinement data.
For the experiments related to the class \textit{mountain bike}, we only retained samples not containing a frame. This is easily automated by checking multiple pixels along the border for agreement with the gray value of the artifact.  
As it is challenging to automatically filter out images containing the \textit{carton}-related watermarks, we did so manually for the corresponding experiments and permitted any correctly classified image without the spurious feature to be used for refinement. 
For the ISIC dataset, we manually removed samples containing the colored patches from the data after training.
\begin{figure}[t]
    \centering
    \includegraphics[width=.25\linewidth]{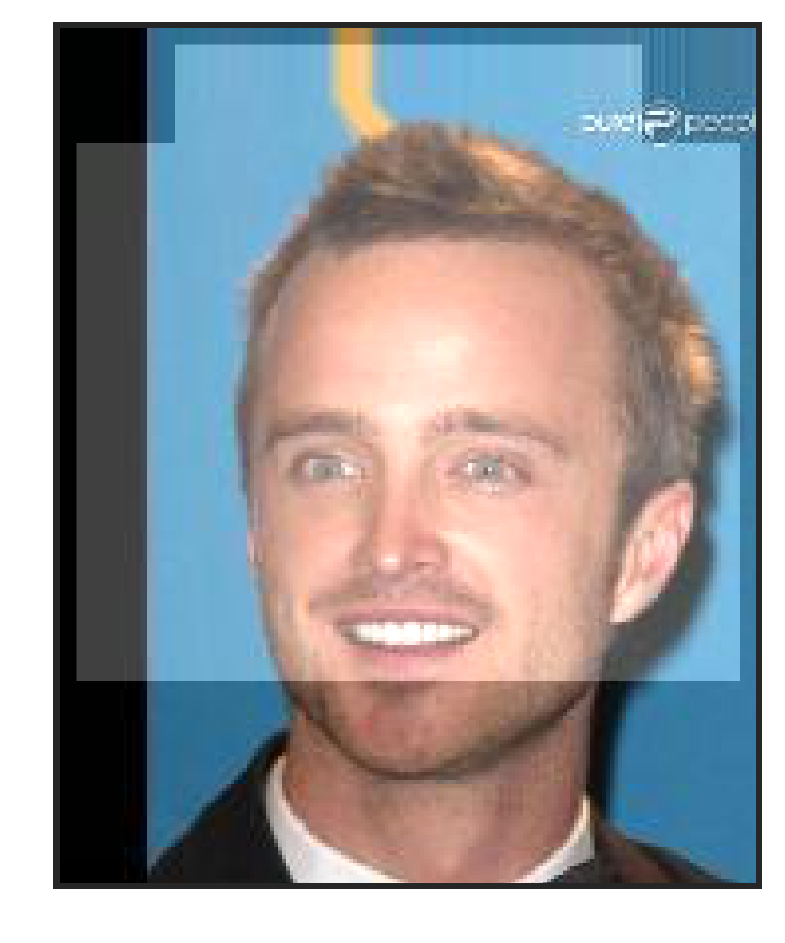}
    \caption{Example of an image from the CelebA dataset. The highlighted area is the mask within which relevance must be concentrated to be accepted as correctly explained.}
    \label{app:fig:celeba:mask}
\end{figure}
For the experiments on CelebA in Section~5 of the main paper, as the task was more open and no specific Clever Hans feature was targeted, we simulated user-verification by checking for agreement of an LRP-based explanation~\cite{Bach2015,Montavon2019a} of the model's decision with an expected location of hair features. To this end we defined a mask, seen in Fig.~\ref{app:fig:celeba:mask}, which roughly corresponds to where we expect features that a user would use to judge a celebrity's hair color to lie. We considered samples to be user-verified if at least 75\% of the absolute LRP relevance for the correct class lies within this mask.

\subsection{Task Selection and Preprocessing}
We will first describe the choice of classes the ImageNet experiments are based on, and then provide further information about the way the individual test samples for the different poisoning scenarios have been manipulated.

The classes to be discriminated were selected based on their similarity to the target classes for which artifacts were identified. Similarity was measured by the mean absolute difference of the soft-max output (see Table~\ref{app:tab:dist}). The selection was done using all training samples of the respective target class. We adopted this selection scheme because natural spurious signals cannot be expected to have a strong enough effect to facilitate significant confusion between classes that are only weakly related.

\begin{table}[t]
    \centering
    \begin{tabular}{|c|l|l|c|}
     \hline
     ID & Class Name & Target & Distance \\\hline
 519 & crate & carton & 0.685 \\
 549 & envelope & carton & 0.691 \\
692 & packet & carton & 0.695 \\
 444 & bicycle-built-for-two & mountain bike & 0.834 \\\hline
    \end{tabular}
    \caption{Mean absolute distance to the target class, calculated from the soft-max outputs of a pretrained ResNet50 network.}
    \label{app:tab:dist}
\end{table}

In the following we will describe the process of manipulating individual data points to shift the distribution of spurious artifacts and create the poisoned dataset for evaluation. The poisoning level denotes with what probability any data point in the clean evaluation dataset is manipulated.

For the class \textit{mountain bike}, we resize the original image to fit into the frame-shaped artifact. The modified image consists of the frame and the shrunken image. We prefer the resizing approach over simply pasting the frame onto the image, as it avoids to remove potential class evidence near the border of the image. Furthermore, we removed the one mountain-bike image from the test set that exhibits the frame artifact.

For the \textit{carton} class, we paste the watermark and URL occurring on images of this class to the center and bottom right, respectively, of the image to be modified. In particular, we manually recreated the URL artifact to look like the ones found in the dataset, but without any additional background. Similarly, the watermark artifact we used was cleaned of its background and pasted with transparency.

Due to the small number of evaluation samples in the ImageNet dataset (50 per class, i.e. the official validation set), we apply data augmentation in order to get robuster estimates. We do this by adding a horizontally flipped version of each samples to the test set.

\subsection{Refinement Hyper-parameters}
In this section we provide the hyper-parameter space we searched over for each refinement method, as well as some additional technical details.

\paragraph{Retraining}
For the Retrain baseline, all layers of the pretrained models are fine-tuned on the available refinement data using the Adam optimizer~\cite{Kingma2015Adam} without further regularization. We use 20\% of the refinement data (rounded up) as validation set and optimize over the number of epochs $N_e \in \{ 1,5,10,20,30,50,100\}$ for all datasets. The learning rates are set as follows. MNIST: $1\cdot10^{-3}$, ImageNet(ResNet50): $5\cdot10^{-6}$, ImageNet(VGG-16): $5\cdot10^{-5}$, ISIC: $1\cdot10^{-7}$. 
We fine-tuned the networks with frozen batch-norm parameters, and gradients clipped to $10^{-3}$.

\paragraph{RGEM and Ridge}
We optimize over the regularization parameter $\lambda \in \{$0.0001, 0.001, 0.01, 0.1, 1, 10, 100, 1000, 10000$\}$. The difference between the baselines is that Ridge regresses the true labels $y$ of the refinement samples whereas REGEM regresses the outputs $\hat{y}$ of the original model.

\paragraph{EGEM and PCA-EGEM}
We optimize over the refinement parameter $\alpha \in \{$0.00001, 0.0001, 0.001, 0.01, 0.1, 0.2, 0.3, 0.4, 0.5, 0.6, 0.7, 0.8, 0.9, 1$\}$. The activations to be refined are the ones immediately after every ResNet50 or VGG-16 block for the parts of the models that are derived from those architectures and additionally after every ReLU activation following a fully connected or convolutional layer outside of such blocks.

\section{Varying Slack}
\label{app:slack}
\begin{figure*}[t]
    \centering
    \includegraphics[width=.85\textwidth]{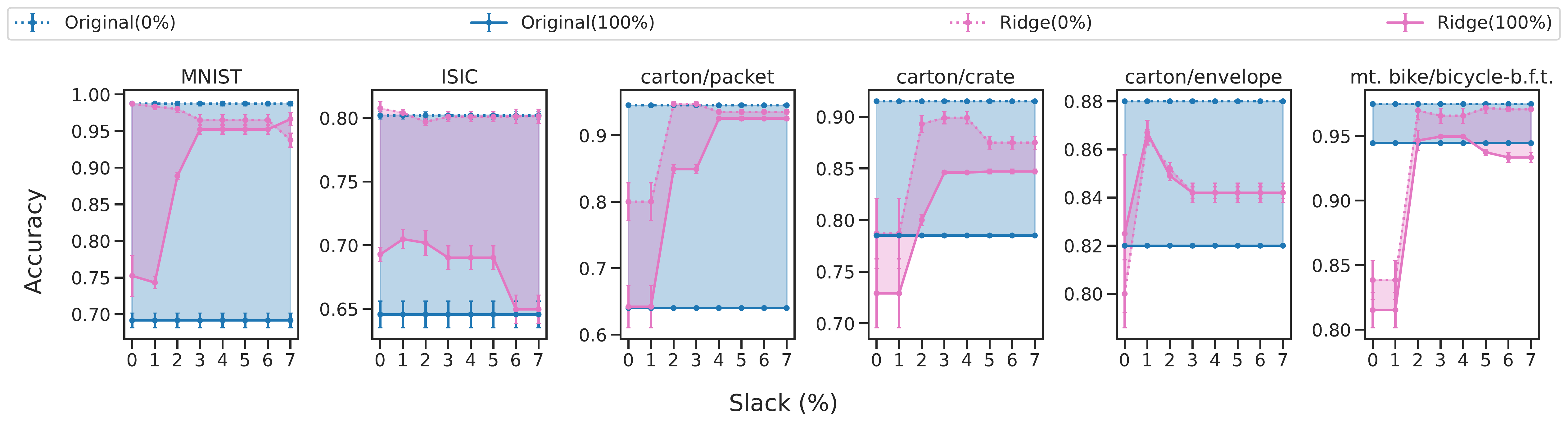}
    \includegraphics[width=.85\textwidth]{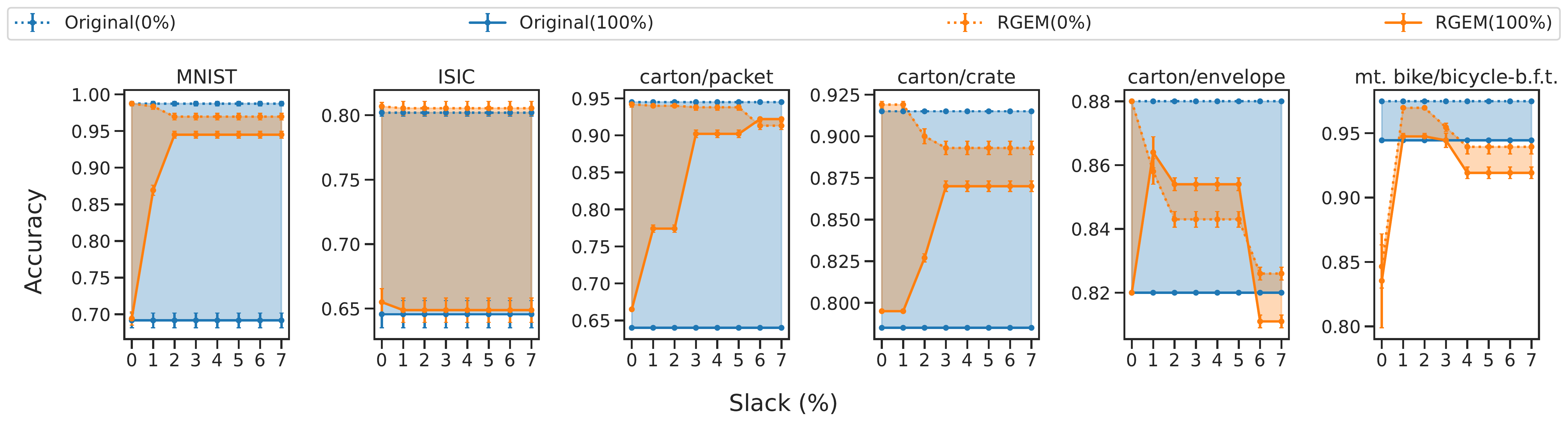}
    \includegraphics[width=.85\textwidth]{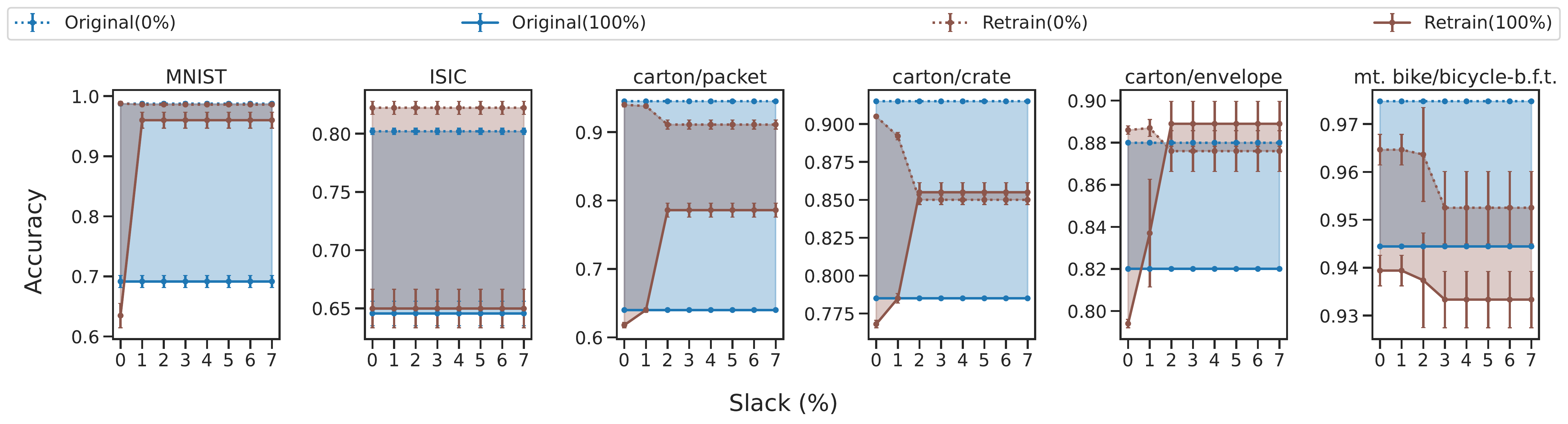}
    \includegraphics[width=.85\textwidth]{figures/NETPCA_vary_slack.pdf}
    \includegraphics[width=.85\textwidth]{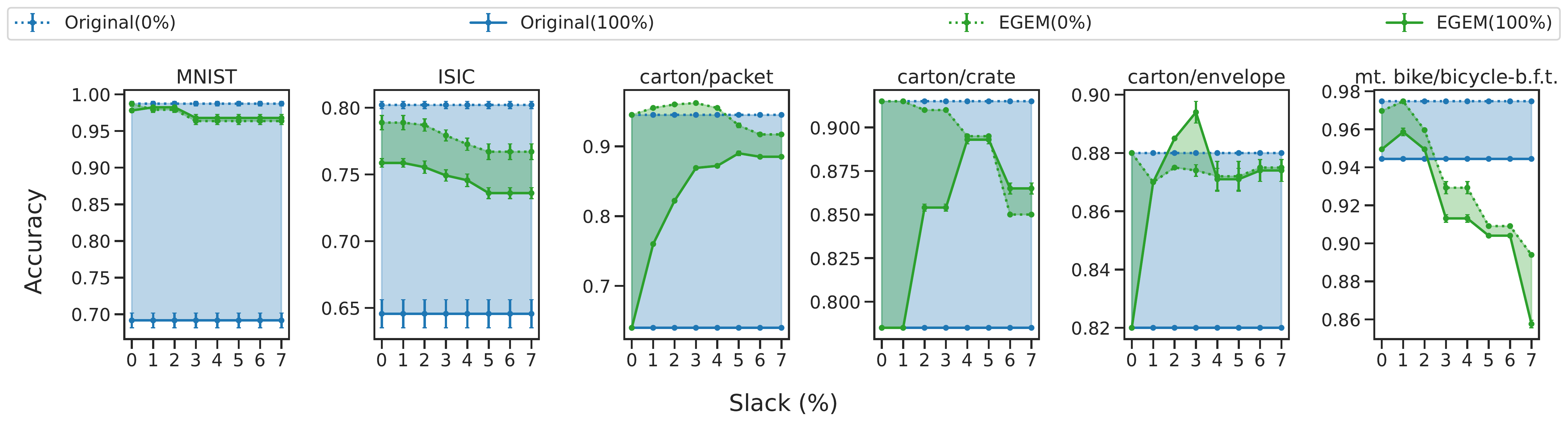}
    \caption{Accuracy under variations of the slack parameter for each refinement method. Higher slack means higher refinement-data loss is accepted when selecting the refinement hyper-parameter.}
    \label{app:fig:eval:slack}
\end{figure*}

\begin{figure*}[t]
    \centering
    \includegraphics[width=\textwidth]{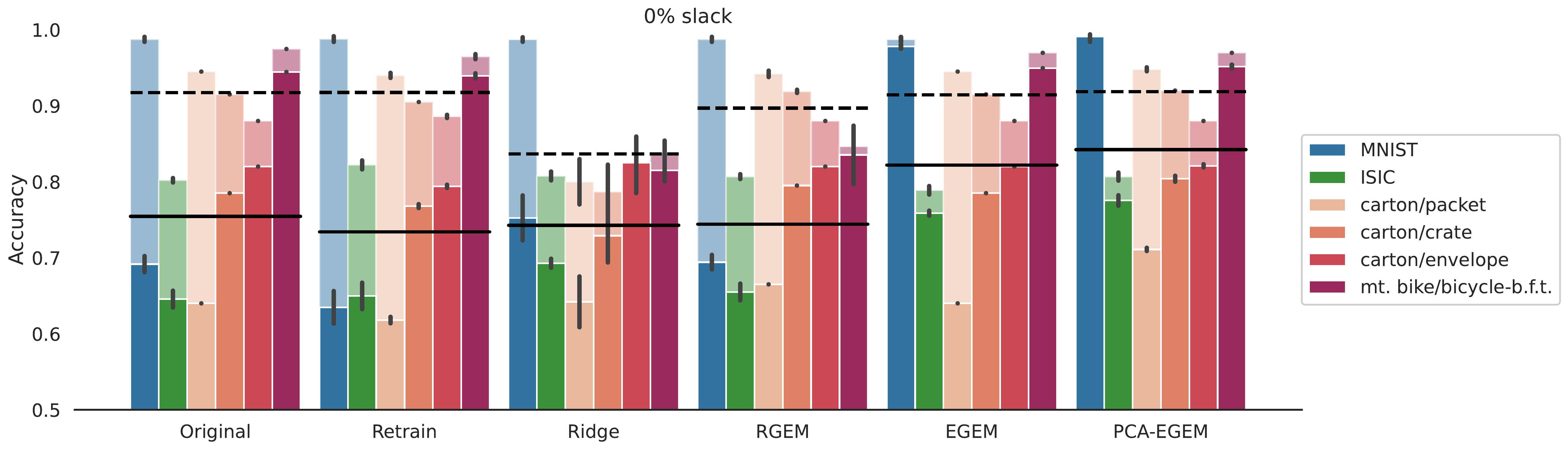}
    \includegraphics[width=\textwidth]{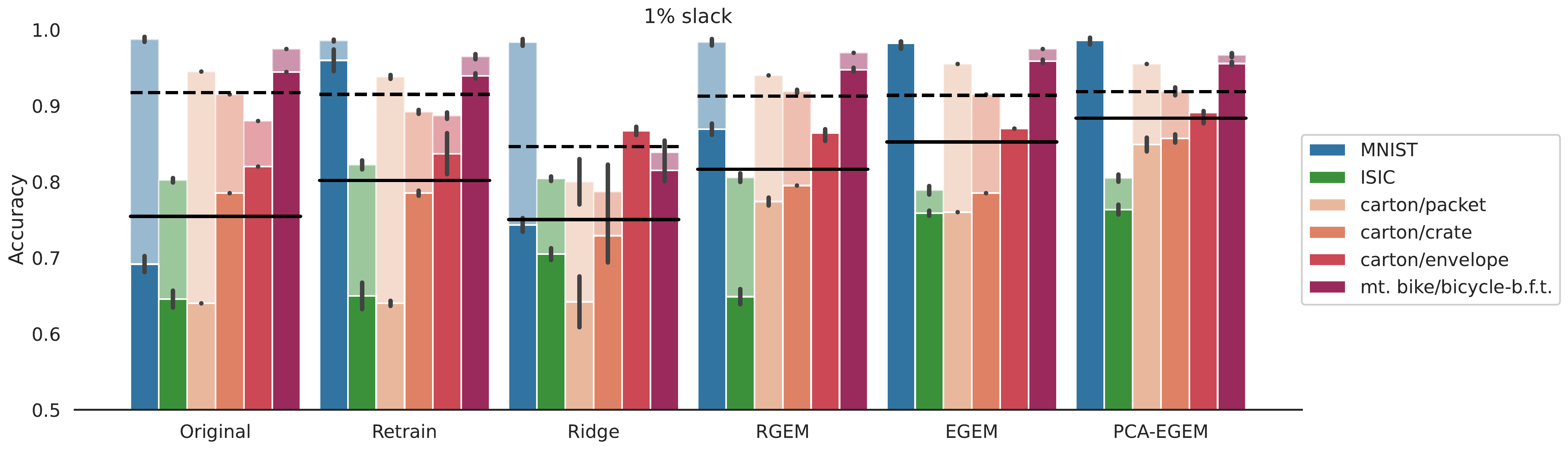}
    \includegraphics[width=\textwidth]{figures/all_vary_d_bars_slk0.05.pdf}
    \includegraphics[width=\textwidth]{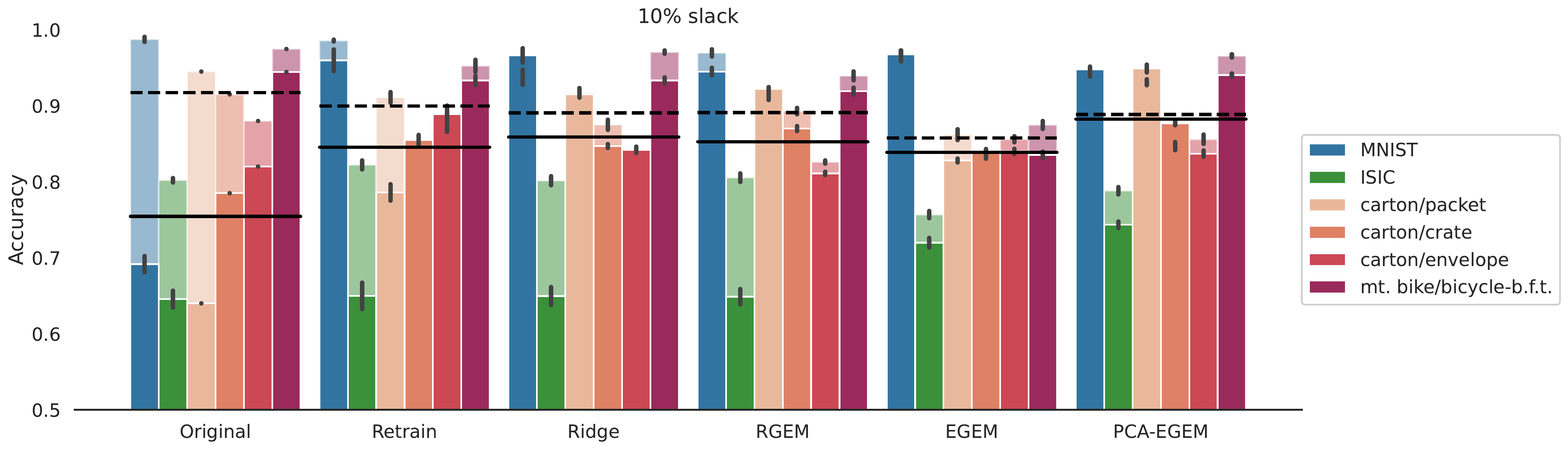}
    \caption{The accuracy for uniform 0\% (lighter shade) and 100\% (darker shade) poisoning with the spurious artifact under variations of the slack parameter. Solid lines show average 100\%-poisoned accuracy and dashed lines show average clean-data accuracy over all datasets. The results shown are the mean accuracy and standard deviation obtained using 700 refinement samples per class on the respective test sets over five runs.}
    \label{app:fig:eval:slack-heaps}
\end{figure*}

In this section we report the 0\%-poisoning and 100\%-poisoning results for varying the value of the slack variable in the hyper-parameter selection for all models in Fig.~\ref{app:fig:eval:slack}. Low slack means that we require the clean-data accuracy of the refined model to be close to the unrefined model, where larger slack allows for increasingly stronger deviations from this base accuracy. Larger slack implies stronger refinement.
Additionally, we show the progression of the average performance with increasing slack in Fig.~\ref{app:fig:eval:slack-heaps}.
The $5\%$ slack level chosen in the main body of the paper can be taken as a rule of thumb, which appears to lead to effective refinement without sacrificing too much clean-data accuracy in most cases. Yet, it can be seen that the optimal level depends on the method and the dataset. For example, at $5\%$ slack PCA-EGEM loses about $4\%$ accuracy on clean and poisoned MNIST data, whereas it would be able to perfectly refine the model without loss of accuracy at the $0\%$ level. 

It can also be observed in Fig.~\ref{app:fig:eval:slack} that the optimal slack value on the various ImageNet tasks is not uniform, which is surprising, since all tasks containing the `carton' class are poisoned with the same artifact. We assume that this effect arises from the other class losing critical features at different refinement levels.
Fig.~\ref{app:fig:eval:slack} demonstrates that PCA-EGEM leads to the most effective refinement over a large number of slack values -- the advantage over other methods being even more apparent in settings with lower slack values than the 5\% presented in the main body of the paper.

\section{Varying Samples}
\label{app:samples}
\begin{figure*}[t]
    \centering
    \includegraphics[width=.85\textwidth]{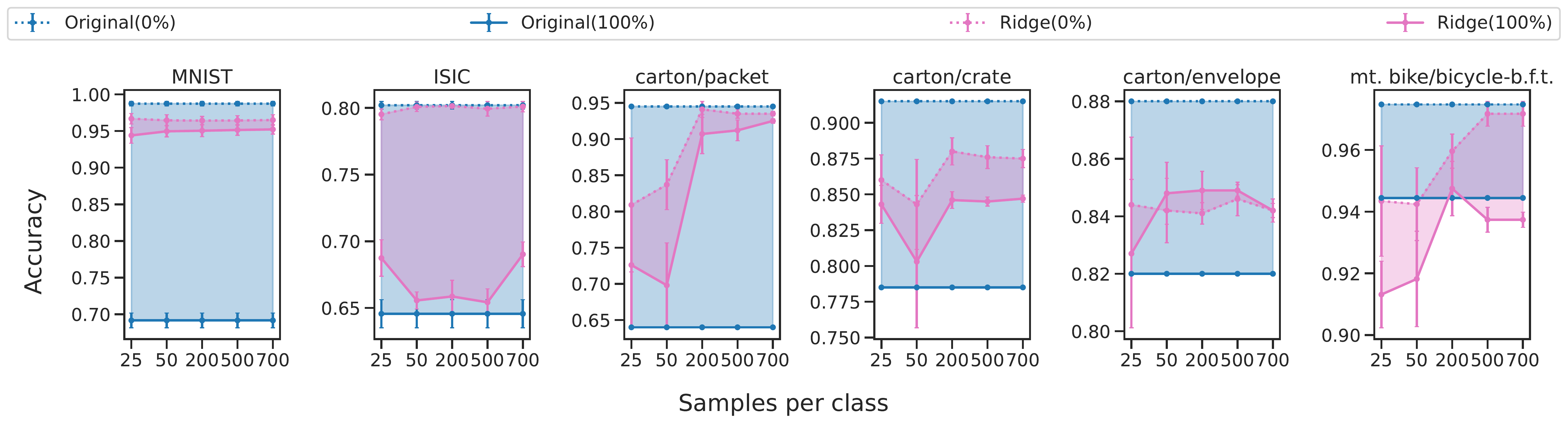}
    \includegraphics[width=.85\textwidth]{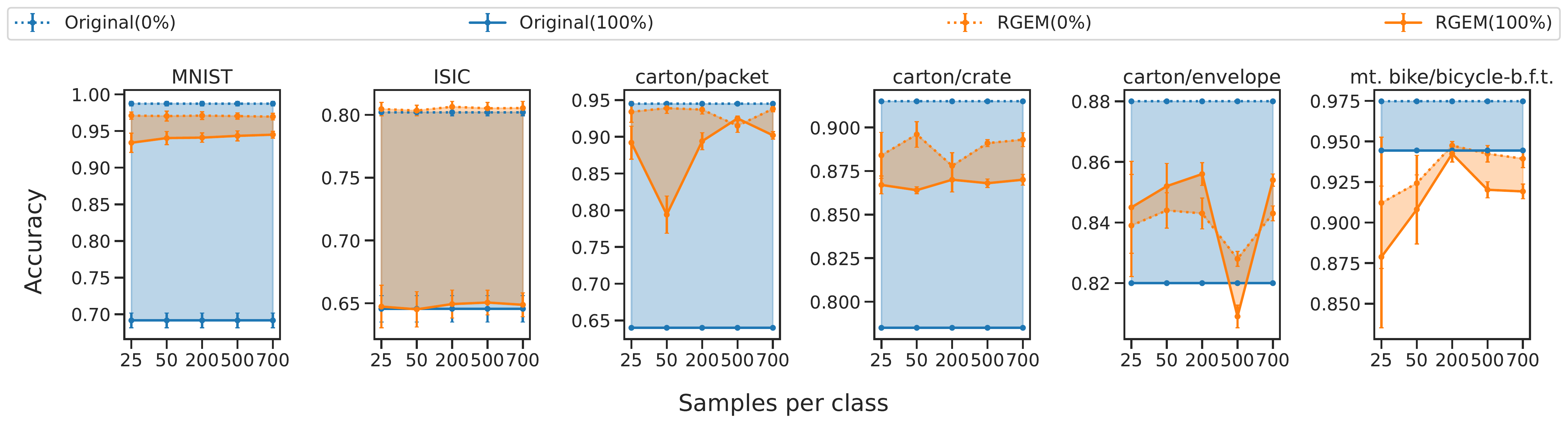}
    \includegraphics[width=.85\textwidth]{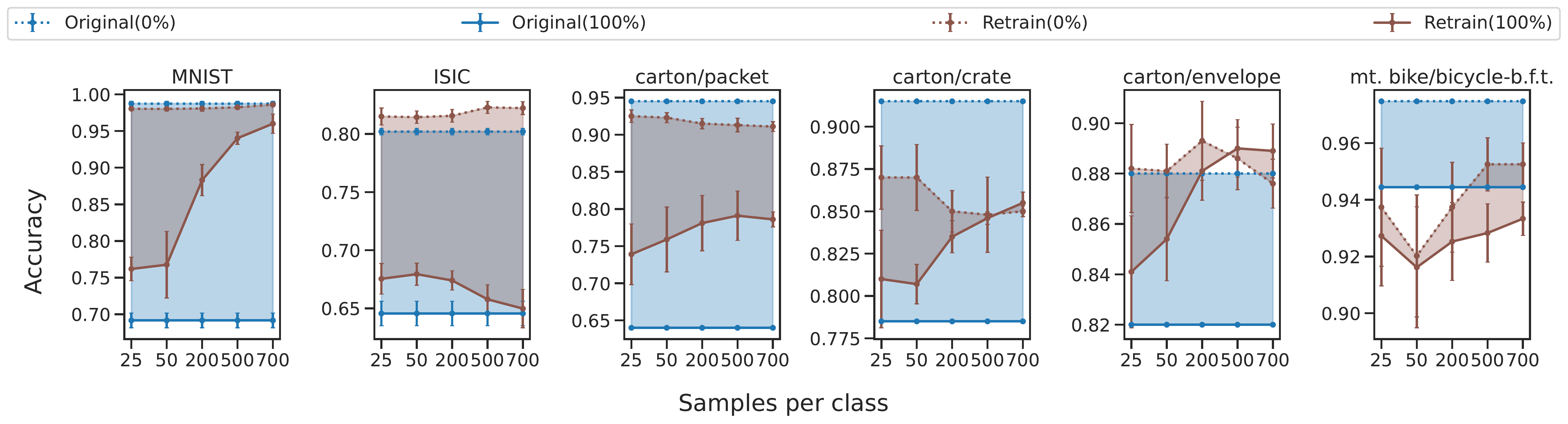}
    \includegraphics[width=.85\textwidth]{figures/NETPCA_vary_samples.pdf}
    \includegraphics[width=.85\textwidth]{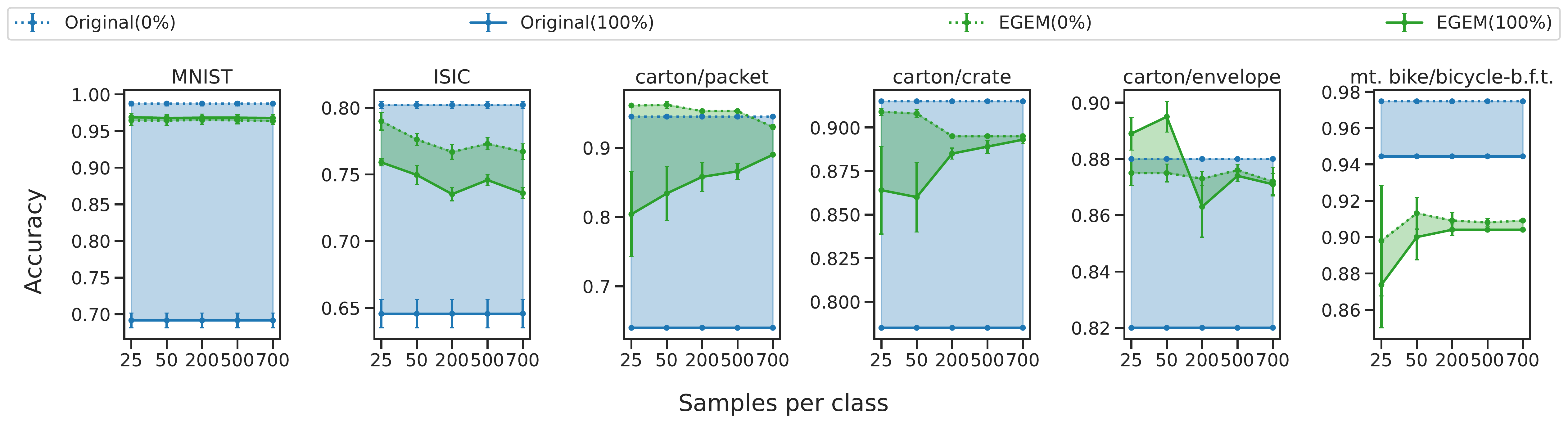}
    \caption{Accuracy under variations of the number of available samples for each refinement method.}
    \label{app:fig:eval:samples}
\end{figure*}

In this section we report the 0\%-poisoning and 100\%-poisoning results for varying the number of examples available for refinement and hyper-parameter selection for all models in Fig.~\ref{app:fig:eval:samples}. 

\begin{figure*}[t!]
    \centering
    \includegraphics[width=.85\textwidth]{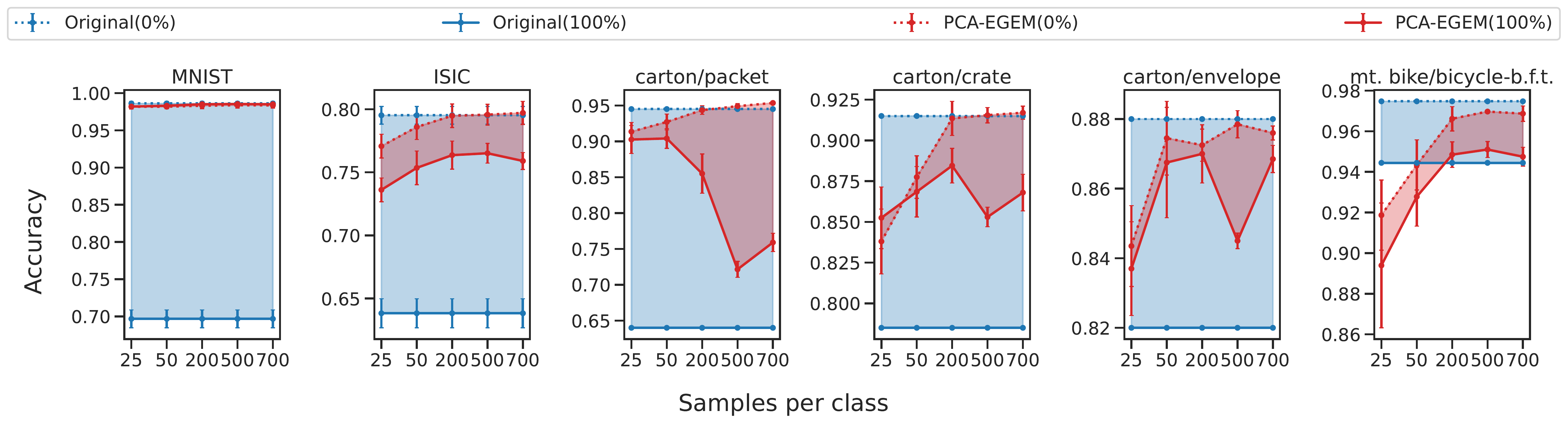}
    \caption{
Effect of PCA-EGEM refinement on accuracy under variations of the number of available samples for each dataset. Slack is set to 1\%.}
    \label{app:fig:eval:samples_pca_s1}
\end{figure*}

Additionally, in Fig.~\ref{app:fig:eval:samples_pca_s1} we report the same for 1\% rather than 5\% slack for PCA-EGEM, since this is the lower limit of our suggested range. While the 100\%-poisoned accuracy does not change by much on average, it can be seen that the clean-data accuracy often converges with fewer samples than for 5\% slack and that for the experiments including the `carton' class, 100\%-poisoned accuracy suffers from weaker refinement. We hypothesize that the drop of poisoned accuracy, in particular for the `carton/packet' task, stems from a misalignment of the neural basis with the CH feature. Thus, samples that contain features similar to the CH feature activate similar latent dimensions and are by this way not completely disentangled from the CH feature. This, in turn, leads to weaker pruning of these shared dimensions if such samples are present in the refinement data.

\section{Additional Experiments on CelebA}
\label{app:exp:celeba}
In this section, we provide further analysis of the effect of PCA-EGEM on the CelebA dataset~\cite{LiuLWT2015} and report additional results.

\subsection{Effect of PCA-EGEM on Precision and Recall}
\begin{figure*}[htb!]
    \includegraphics[width=.9\linewidth]{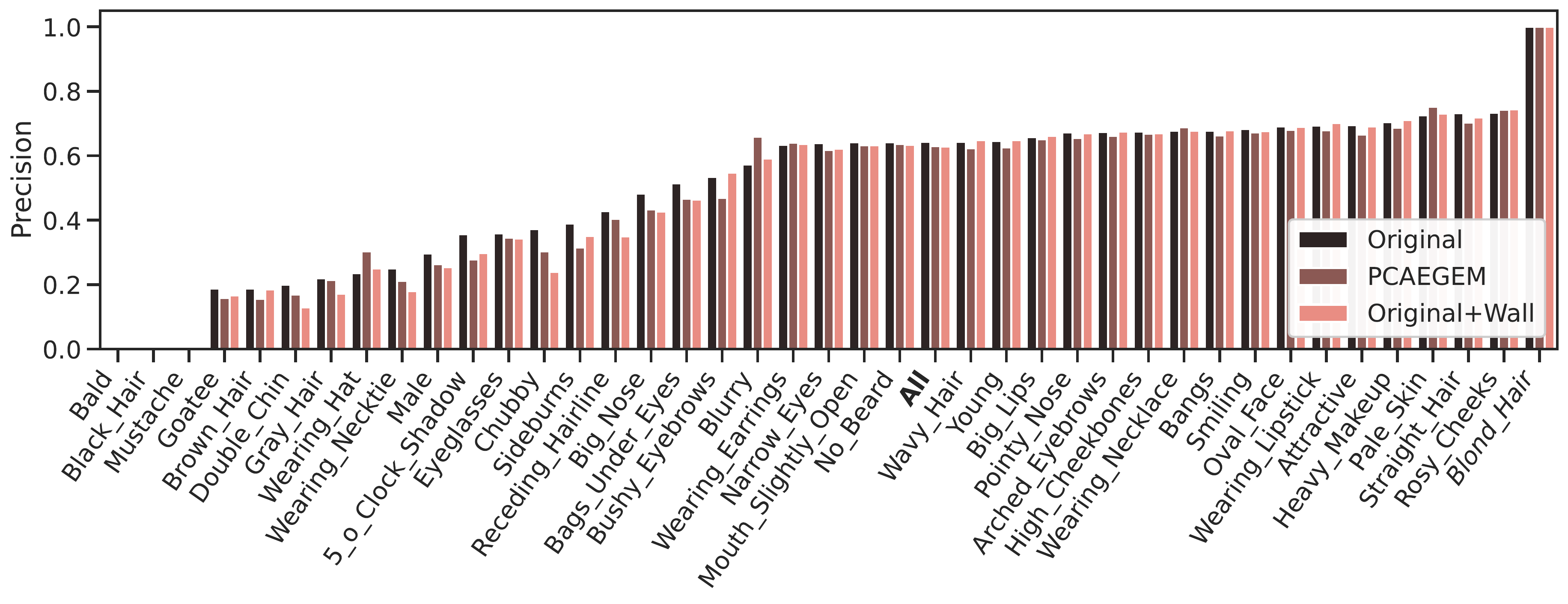}
    \includegraphics[width=.9\linewidth]{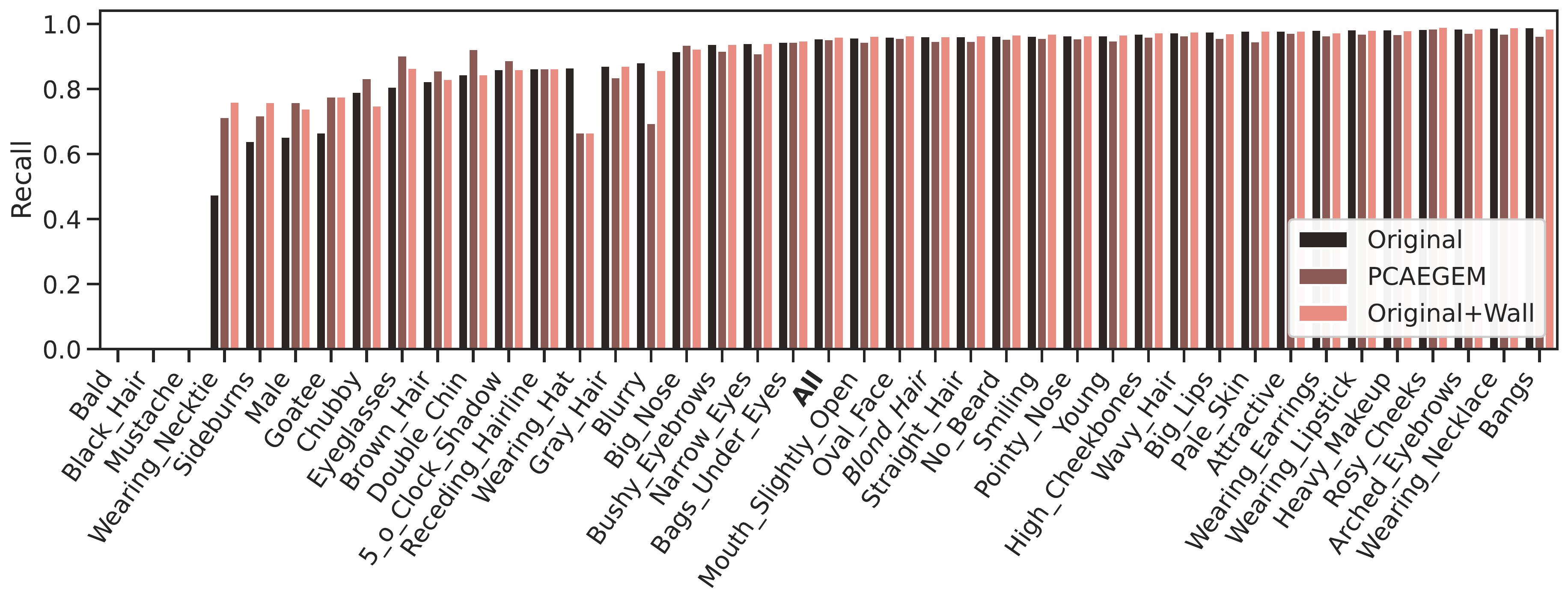}
\caption{Comparison of precision and recall of the pretrained model before (Original) and after refinement (PCA-EGEM). The metrics are calculated on subsets of CelebA containing only samples exhibiting the attribute on the x-axis. `All' is sampled from the whole test set.} \label{app:fig:celeba:results}
\end{figure*}

As discussed in the main paper, application of PCA-EGEM to CelebA data tends to harmonize recall of the attribute to be detected (``Blond Hair'') across subgroups (cf.\ Fig.\ \ref{app:fig:celeba:results}). In particular, recall for subgroups where the attribute to be detected has low-prevalence tends to be increased. However, two subgroups, `Wearing\_Hat' and `Blurry', go contrary to this trend. Specifically, their recall is substantially reduced after the application of PCA-EGEM. We give the following a detailed explanation for this effect.

In both the ``Wearing Hat'' and ``Blurry'' subgroups, the overall evidence for the attribute ``Blond Hair'' is inherently weak. This is caused either by occlusion by the hat or obscuring by the blur. To overcome this loss of information, i.e.\ to achieve high classification accuracy, the original model has likely in such cases to found alternatives to actual hair features, in particular, by incorporating additional \textit{weakly correlated} features into the decision strategy. Examples of weakly correlated (CH-type) features include wearing lipstick or heavy makeup (cf.\ Figure~\ref{app:fig:data:celeba:correlation}). Such modification of the decision strategy towards CH features is however only desirable from the perspective of maximizing classification accuracy for instances where hair features are not easily accessible (i.e.\ mostly instances from the subgroups ``Wearing Hat'' and ``Blurry''). Because of the rarity of such instances, and consequently the low probability that they are contained in the limited available data, the CH-based decision strategy the model applies to them is likely to be pruned away by PCA-EGEM. As a result, instances from the ``Wearing Hat'' and ``Blurry'' subgroups that were already on the borderline of evidential support for blond hair may no longer be recalled. More generally, a drop in recall after the application of PCA-EGEM may indicate that the original model has made use of weakly correlated CH-type features instead of true well-generalizing ones for its prediction task.

Interestingly, on the same subgroups ``Wearing Hat'' and ``Blurry'', we can observe in Fig.~\ref{app:fig:celeba:results} an increase of precision that mirrors the decrease of recall. To explain this, a similar argument can be made: Because PCA-EGEM prunes weakly correlated CH features from the model, the few instances of those subgroups where the hair is clearly detectable by the model have relatively higher detection scores, compared to instances where the hair is barely visible. This makes it more likely that the top-few instances predicted after the application of PCA-EGEM to have the attribute ``Blond Hair'' truly have the attribute ``Blond Hair'', thereby leading to a higher precision score.

More generally, we see from Fig.~\ref{app:fig:celeba:results} an inverse correlation between the effect of PCA-EGEM on precision and on recall. Specifically, an increase in recall often results in a decrease in precision and vice-versa. This is consistent with the commonly made observation of a tradeoff between precision and recall~\citep{buckland1994}.

\bigskip

\subsection{Validation of the Collar Clever Hans Feature on CelebA} 
\begin{figure}[htb!]
    \centering
    \includegraphics[width=.9\linewidth]{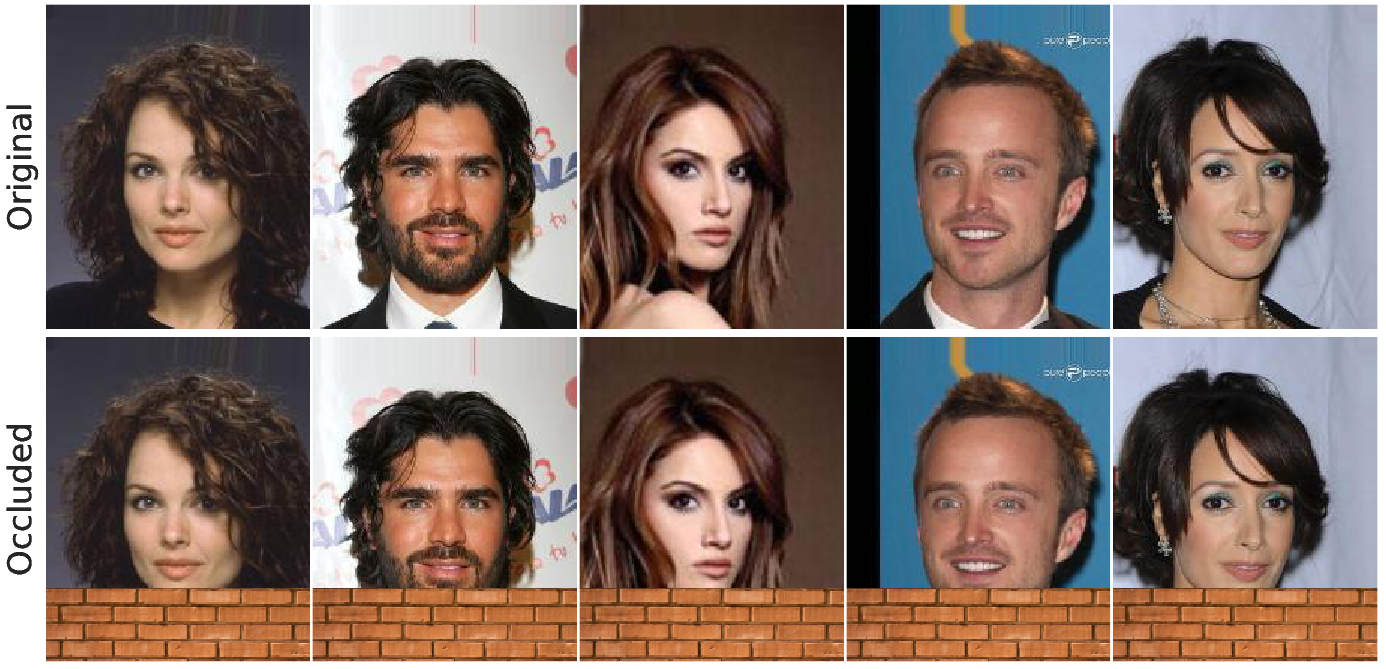}
    \caption{Examples of images drawn from the CelebA dataset before and after occlusion.}
    \label{app:fig:celeba:wall}
\end{figure}

To corroborate our observation that shirt collars are a CH feature, we created a counterfactual dataset where the bottom section of the images is occluded by an image of a wall (see Fig.~\ref{app:fig:celeba:wall}) and repeated our analysis with occlusion instead of refinement. As expected, we can observe that occlusion of the bottom section of the images improves recall of the `Wearing\_Necktie', `Sideburns', `Goatee', and `Male' groups. Furthermore, the recall of the `Wearing\_Hat' group is reduced -- this is most likely due to the fact that for some samples the blond part of the hair would only be visible in the now-occluded part of the image, as it is covered by a hat in the upper parts of the image. 
Overall, these results support our observation of collars being inhibitive CH features.

\section{Additional Experiments on MNIST/ISIC/ImageNet}
\label{app:exp:misc}
In this section, we present additional results to elucidate the differences of the evaluated refinement strategies.

\subsection{Separability of Clean and Manipulated Samples Throughout the Network}
Here we randomly sample $N=10$ instances from the training dataset and measure the separability of these instances from their poisoned counterparts by the $R^2$-score, as defined by Kornblith et al.~\citep{Kornblith2021}:
\begin{align}
\label{app:eq:separability}
    R^2 &= 1-\Bar{d}_{within}/\Bar{d}_{total}\\
    \Bar{d}_{within} &= \sum_{k=0}^1\sum_{m=1}^N\sum_{n=1}^N \frac{cos(x_{k,m}, x_{k,n})}{2N^2}\\
    \Bar{d}_{total} &= \sum_{j=0}^1\sum_{k=0}^1\sum_{m=1}^N\sum_{n=1}^N \frac{cos(x_{j,m}, x_{k,n})}{4N^2}
\end{align}
where the the first subscript ($j$ and $k$) indicates poisoning with the CH feature and $cos(\cdot)$ denotes the cosine distance. 
The score is calculated at the outputs of all layers of the network, or, for VGG-16 and ResNet50 architectures, after each block. 

\begin{figure}[htb!]
    \centering
    \includegraphics[width=.9\linewidth]{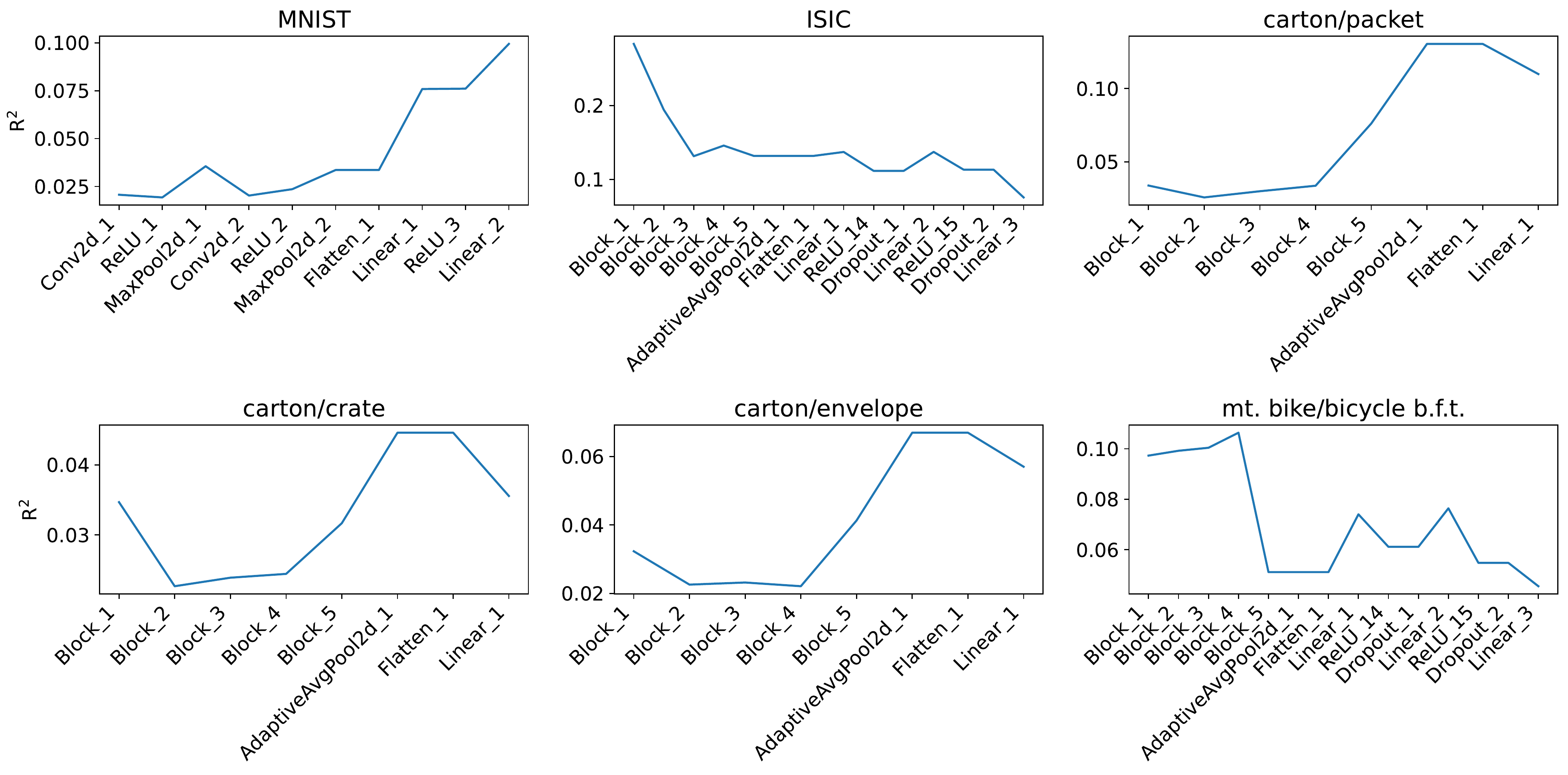}
    \caption{Per-layer separability of clean versus poisoned samples, as measured by Eq.~\ref{app:eq:separability}.}
    \label{app:fig:separability}
\end{figure}

Figure~\ref{app:fig:separability} shows that, in particular for the ISIC and the `mountain bike' experiment, the two groups can be more easily separated at early layers, whereas for MNIST, the CH features seem most clearly expressed just before the last layer. This observation suggests that in some cases, deeper refinement methods, such as (PCA-)EGEM or retraining, may be better suited to remove CH features than methods modifying only the last layer.

\subsection{Change in Output Logits}
\begin{figure}[htb!]
    \centering
    \includegraphics[width=.9\linewidth]{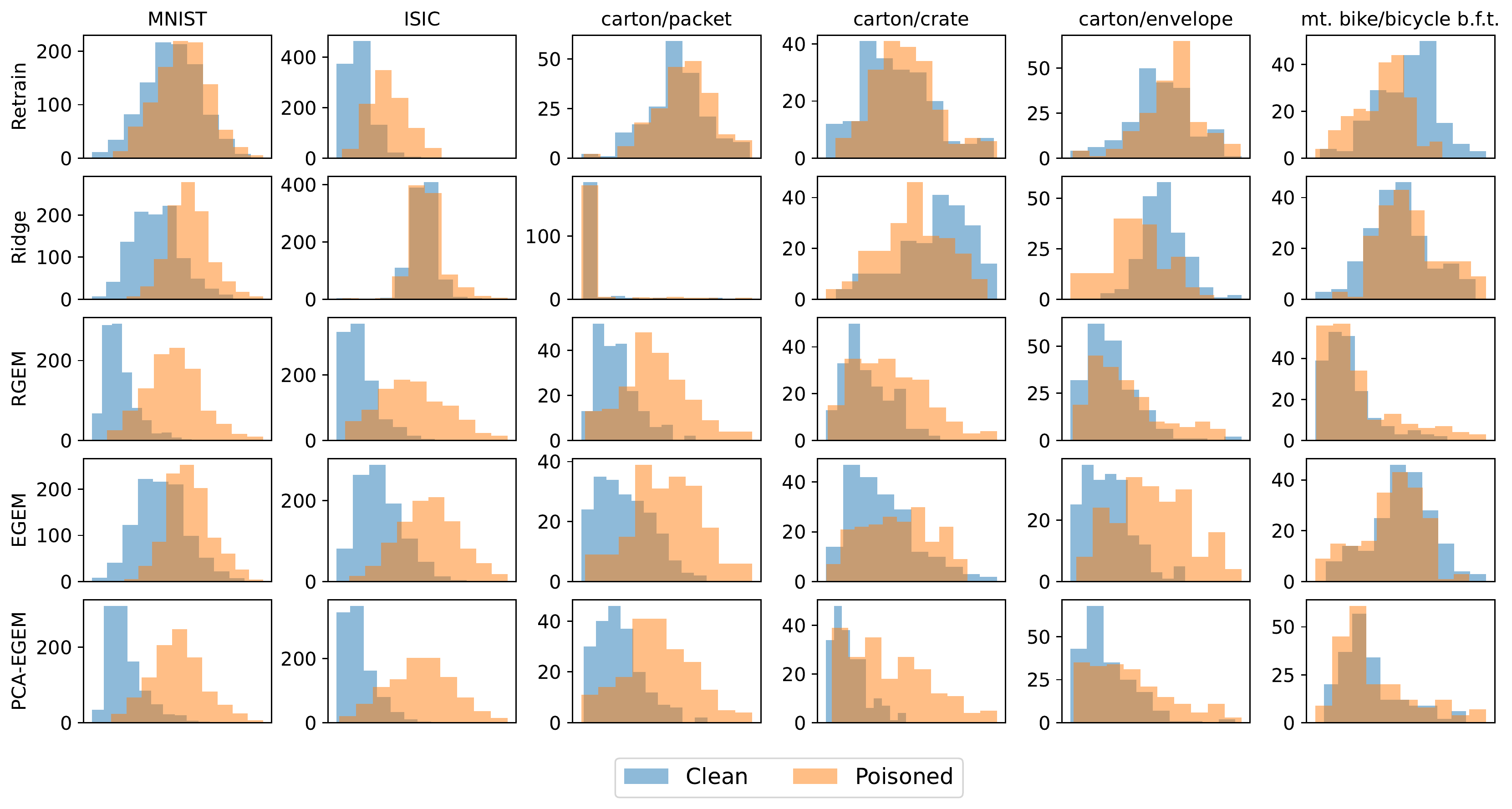}
    \caption{Output logits after refinement for clean and 100\%-poisoned test samples.}
    \label{app:fig:logits}
\end{figure}

For the task of refinement, the ideal outcome would be that the model's response to inputs remains unchanged, except for the samples containing CH features, the sensitivity to which should be refined away. In practice, this is not achieved. Yet, it can be seen in Fig.~\ref{app:fig:logits} that exposure-minimization methods lead to comparatively small changes in the output logits on clean samples and often distinctly larger changes in the poisoned samples when compared to other baselines. We see this as an indication of such methods being more targeted in their change of the model, leaving much of its original functional structure intact.

\putbib
\end{bibunit}